\documentclass[preprint,12pt]{elsarticle}

\usepackage{amssymb}
\usepackage{amsmath}

\usepackage{todonotes}

\journal{Information Fusion}

\usepackage[
  colorlinks=true,
  linkcolor=red,   
  citecolor=green  
]{hyperref}

\usepackage{array}
\usepackage{booktabs}
\usepackage{float} 
\usepackage{placeins} 
\usepackage{longtable} 
\usepackage{adjustbox}
\usepackage{makecell}

\newcolumntype{C}[1]{>{\centering\arraybackslash}m{#1}}
\newcolumntype{L}{@{}l@{}}  

\begin{document}

\begin{frontmatter}



\title{Exploring Embodied Multimodal Large Models: Development, Datasets, and Future Directions}


\author[label1,label2]{Shoubin Chen}
\ead{shoubin.chen@whu.edu.cn}
\author[label1]{Zehao Wu}
\ead{wuzehao@gml.ac.cn}

\author[label1,label3]{Kai Zhang}
\ead{kai.zhang.2021.54@ensta.fr}

\author[label1,label4]{Chunyu Li \corref{cor1}}
\ead{chunyu2020@email.szu.edu.cn}

\author[label1]{Baiyang Zhang}
\ead{zhangbaiyang@gml.ac.cn}
\author[label1]{Fei Ma} 
\ead{mafei@gml.ac.cn}
\author[label1]{Fei Richard Yu}
\ead{richard.yu@carleton.ca}
\author[label1,label2]{Qingquan Li} 
\ead{liqq@szu.edu.cn}

\cortext[cor1]{Corresponding authors.}


\affiliation[label1]{organization={Guangdong Laboratory of Artificial Intelligence and Digital Economy (SZ)},
            city={Shenzhen},
            country={China}}
            
\affiliation[label2]{organization={Guangdong Key Laboratory of Urban Informatics, Shenzhen University},
            city={Shenzhen},
            country={China}}

\affiliation[label3]{organization={ENSTA Paris, Institut Polytechnique de Paris},
            city={Palaiseau},
            postcode={91120}, 
            state={Essonne},
            country={France}}

\affiliation[label4]{organization={Sun Yat-sen University},
            city={Guangzhou},
            postcode={510275}, 
            country={China}}
\begin{abstract}
Embodied multimodal large models (EMLMs) have gained significant attention in recent years due to their potential to bridge the gap between perception, cognition, and action in complex, real-world environments. This comprehensive review explores the development of such models, including Large Language Models (LLMs), Large Vision Models (LVMs), and other models, while also examining other emerging architectures. We discuss the evolution of EMLMs, with a focus on embodied perception, navigation, interaction, and simulation. Furthermore, the review provides a detailed analysis of the datasets used for training and evaluating these models, highlighting the importance of diverse, high-quality data for effective learning. The paper also identifies key challenges faced by EMLMs, including issues of scalability, generalization, and real-time decision-making. Finally, we outline future directions, emphasizing the integration of multimodal sensing, reasoning, and action to advance the development of increasingly autonomous systems. By providing an in-depth analysis of state-of-the-art methods and identifying critical gaps, this paper aims to inspire future advancements in EMLMs and their applications across diverse domains.
\end{abstract}



\end{frontmatter}


\section{Introduction}
\label{sec:intro}

Embodied intelligence, the idea that cognition arises from physical interaction with the environment, emerged as a critique of traditional cognitive theories. Rodney Brooks' 1991 paper \cite{brooks1991intelligence}, ``Intelligence Without Representation,'' argued that intelligent behavior can emerge without relying on internal representations, focusing instead on environmental interaction. This idea was further developed by Varela, Thompson, and Rosch in The Embodied Mind (1991) \cite{varela2017embodied}, which highlighted the role of bodily experience in shaping cognition. Similarly, Lakoff and Johnson's Philosophy in the Flesh (1999) emphasized that cognition is grounded in sensory-motor experiences \cite{lakoff1999review}. The concept also found practical applications in robotics, as seen in the Cog Project, which explored how robots could develop cognition through bodily interaction with the world \cite{brooks1998cog}. Thus, embodied intelligence bridges cognitive science, philosophy, and robotics, offering a more integrated view of mind and body.

With the rapid advancement of large model technology, embodied intelligence is increasingly being integrated with these models. This is a relatively new concept, where researchers seek to apply principles of embodied cognition to large-scale pre-trained models. The aim is to explore how Artificial Intelligence (AI) can develop more flexible and adaptive capabilities through interactions with the environment. The term ``Embodied Multimodal Large Models'' (EMLMs) refers to a class of models that combine multiple modalities of data (e.g., vision, language, and action) with embodied capabilities (e.g., perception and interaction within physical environments). These models are also known by various other names, such as ``Large Embodied Multimodal Models,'' ``Embodied Large Models,'' and ``Large Embodied Models.'' While these terms are often used interchangeably, they all highlight the integration of multimodal understanding with the ability to perceive and interact with the world in a physically embodied manner. In this paper, we adopt the term EMLMs to encompass all such variations. We provide a comprehensive exploration of the development, datasets, and challenges associated with these models, offering a detailed analysis of their current state and future potential.

EMLMs are an exciting and rapidly evolving area within the fields of AI and robotics. Unlike traditional AI systems, EMLMs integrate diverse sensory modalities, such as vision, language, and audio, into agents capable of perceiving and interacting with their physical environments. EMLMs aim to bridge the gap between AI’s abstract reasoning abilities and the real-world complexities, allowing intelligent systems to perceive, act, and learn in ways that are more closely aligned with human cognition. These models can simultaneously process multimodal input and generate outputs that influence the physical world, making them critical for applications such as robotic manipulation, autonomous navigation, human-robot interaction, and immersive virtual environments.

\begin{figure*}[t]
	\centering
	\includegraphics[width=1.0\linewidth, trim={5 65 10 60}, clip]{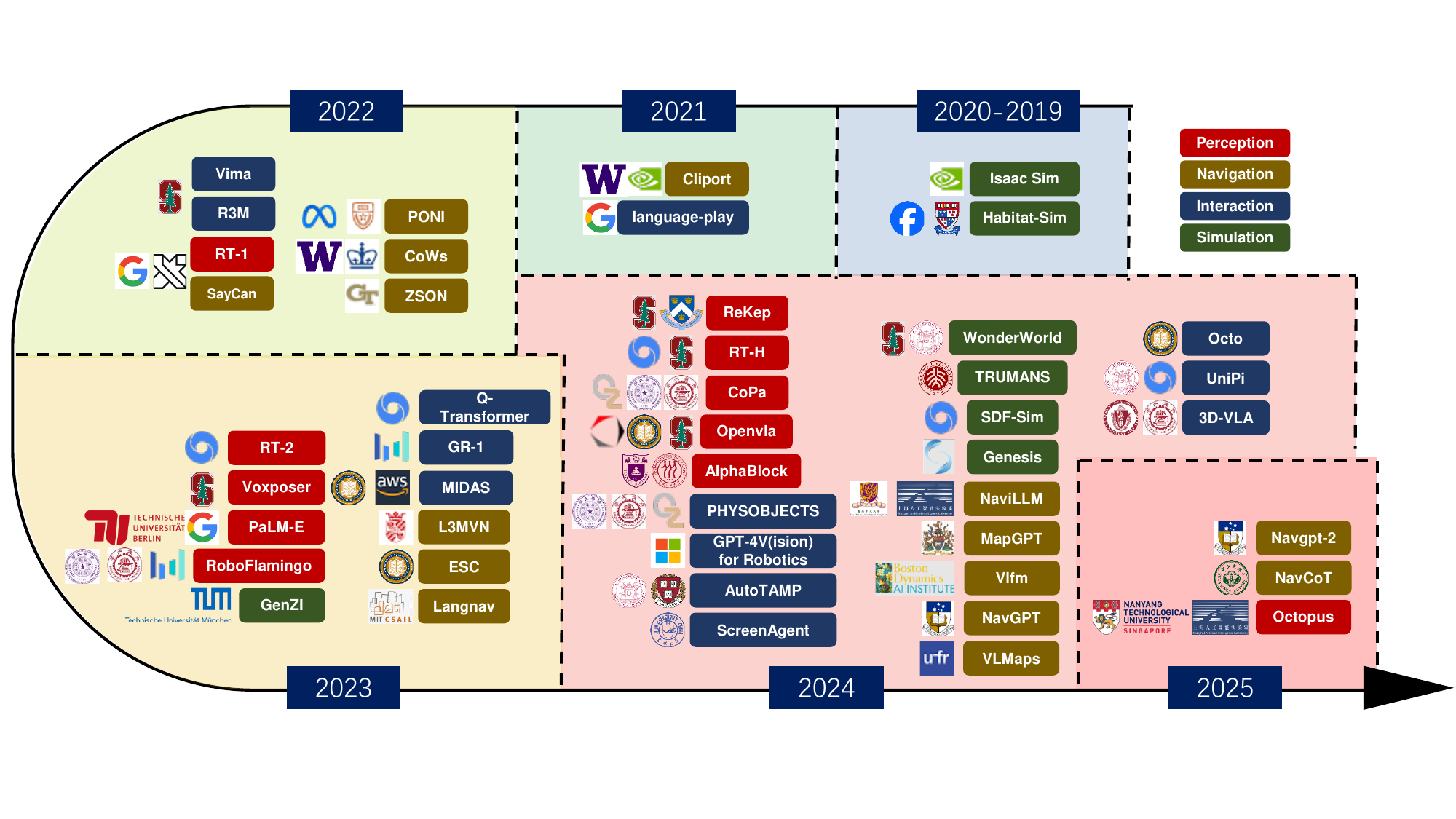}
	
	\caption{A timeline of research progress in the field of Embodied Perception, Navigation and Interaction.}
	\label{fig:Section3}
\end{figure*}

In recent years, the integration of large models with multimodal perception systems, such as embodied agents, has led to the development of breakthrough models capable of tackling increasingly complex tasks. However, the field of embodied intelligence with large models remains in its early stages, and several challenges persist. These include enhancing model scalability and generalization, improving the ability to handle complex tasks, and advancing the capacity of embodied agents to interact more effectively with their environments.

Although significant progress has been made in this field, several key issues persist in the current review papers on EMLMs. First, most existing reviews primarily focus on traditional large models in natural language processing, such as LLMs \cite{zhao2023survey} \cite{chang2024survey} \cite{naveed2023comprehensive}, large vision models, and language-vision models \cite{li2024multimodal}, rather than explaining the integration of embodied agents with large models. Second, even some reviews do focus on this integration, the scope is often too broad. For instance, papers \cite{wu2024embodied} \cite{du2024advancements} \cite{liu2024aligning} \cite{roy2021machine} concentrate on the entire development process of embodied intelligence, including both software and hardware, without delving deeply into the role of multimodal large models in the evolution of embodied intelligence. Additionally, some prior works \cite{firoozi2023foundation} \cite{duan2022survey} \cite{xi2023rise} were published before the most recent rapid advancements in the field, limiting their ability to capture the state-of-the-art developments. Some papers focus solely on specific big model technologies within the embodied intelligence full stack, rather than addressing the big model technologies across each link of the stack. For instance, the paper \cite{ma2024survey} primarily examines the big model in the intelligent agent operation segment, but does not consider the big model in the navigation component.

To address these gaps, this paper provides a comprehensive review of recent developments in EMLMs, focusing on four key areas: (1) technical advancements in foundational large models, such as LLM and LVMs, which are driving the development of EMLMs, (2) the current technical roadmap for EMLMs across various tasks, including perception, navigation, interaction, and simulation, (3) the impact of multimodal datasets on model performance, and (4) challenges and opportunities for the future development of EMLMs. Our aim is to offer a thorough overview of existing progress and identify potential future directions. We hope that this review will serve as a valuable reference and a source of inspiration for researchers in this field.

The main contributions of this paper are summarized as follows:

\begin{enumerate}
    \item First Systematic Review of Research Progress in EMLMs: To the best of our knowledge, this paper is the first to systematically review the research progress in the field of EMLMs, addressing a significant gap in the existing literature.
    \item Comprehensive Full-Stack Analysis: This study analyzes 300 research papers and conducts a comprehensive full-stack analysis of EMLMs, covering basic big models, embodied perception big models, embodied navigation big models, embodied interaction big models, simulation techniques, and datasets. Through in-depth analysis, readers can have a comprehensive understanding of the development of EMLMs.
    \item Benchmark datasets and Collection Methods: We summarize the benchmark datasets used in EMLMs, detailing their collection methods. Additionally, we analyze the key features of these datasets, including data format, functionality, applicable platforms, as well as their advantages and limitations.
    \item Main Findings and Future Research Directions: Finally, we discuss the primary findings of this review, provide insights into the application of EMLMs in embodied agents, and outline promising directions for future research in this rapidly evolving field.
\end{enumerate}

The remainder of this paper is organized as follows: In Section \ref{sec:LM}, we explore the development of large models, and other emerging architectures, outlining their contributions to the EMLMs. Section \ref{sec:EMLM} delves into the development of EMLMs, discussing key components, such as embodied perception, navigation, interaction, and simulation, which are critical for enabling EMLMs to engage with and respond to real-world environments. Section \ref{sec:DATA} provides an overview of the datasets used to train and evaluate EMLMs, highlighting the challenges and importance of high-quality, diverse data for effective learning. In Section \ref{sec:CFD}, we address the major challenges faced by the field, including scalability, generalization, real-time decision-making, and the seamless integration of different modalities. Additionally, we identify promising future research directions. Finally, Section \ref{sec:CON} concludes the paper by summarizing the key findings and offering insights into the future trajectory of EMLMs.


\section{Development of Foundational Models for Embodied Multimodal Large Models}
\label{sec:LM}
The evolution of large models, particularly in natural language processing, computer vision, and deep learning, has paved the way for the emergence of EMLMs. These advanced systems leverage the integration of multiple modalities, such as vision, language, audio and touch, to enable more natural and intuitive interactions with the physical world. The advancements in large-scale pretraining and the scaling of neural networks have facilitated the creation of models capable of processing multimodal data while embodying a deeper understanding of context, actions, and interactions, setting the stage for the next frontier in AI.

In this section, we first review the development of embodied agents, which are the carriers of the large models in the physical or virtual worlds. Then, we introduced the basic techniques of multimodal large models used in EMLMs, including language, vision, audio and touch models.

\subsection{Embodied Agents}
\begin{figure}[t]
	\centering
	\includegraphics[width=1.0\linewidth]{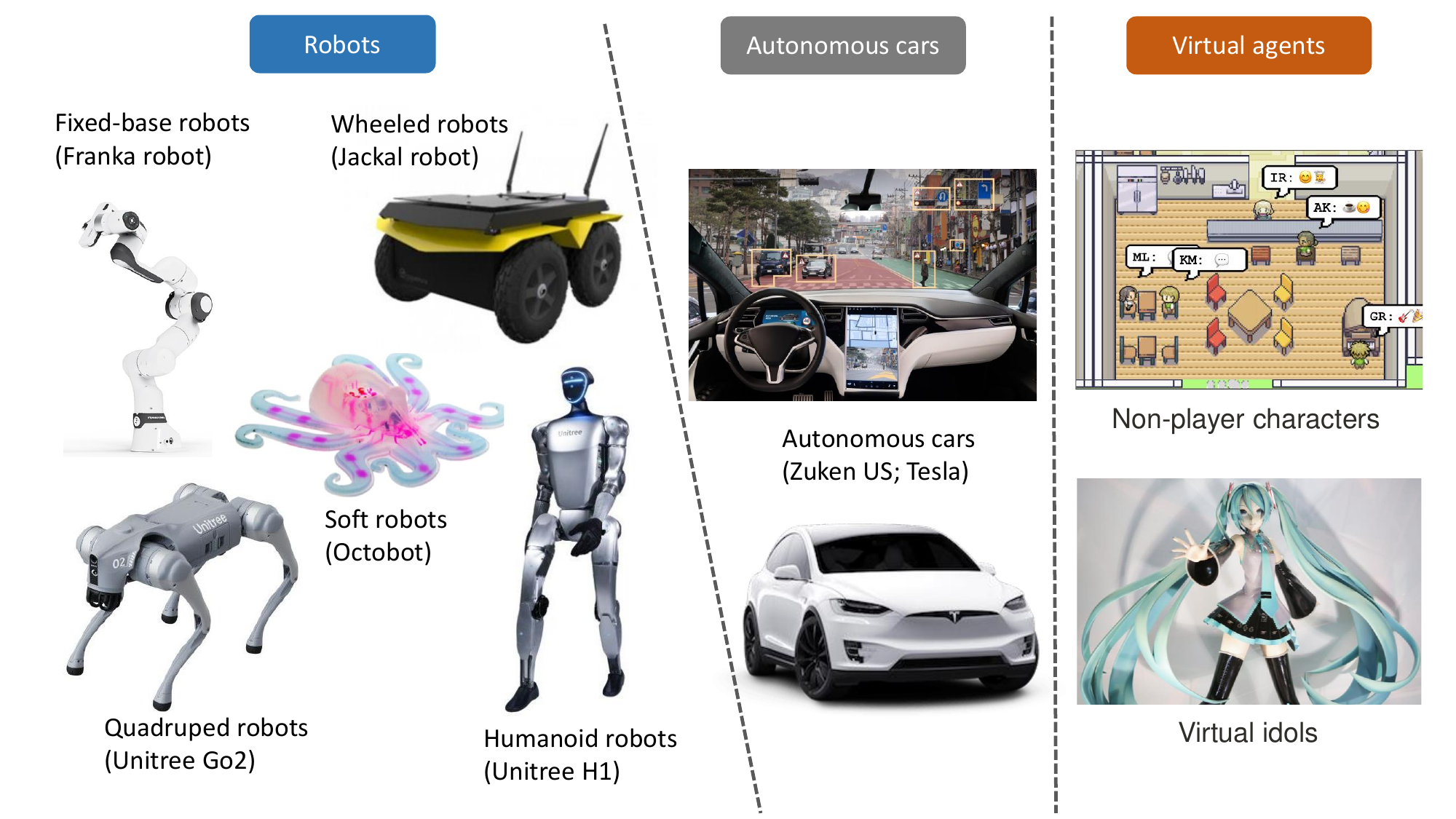}
	
	\caption{Examples of embodied AI agents.}
	\label{fig:agents}
\end{figure}

Embodied agents are autonomous entities that have physical or virtual body and are able to perceive, act and interact with the environments. They have various types, such as the robots, autonomous cars, virtual agents, etc. 

As shown in Fig.~\ref{fig:agents}, robots are the most popular agents used in existing embodied AI algorithms. Depending on the applications, robots have various forms, including fixed-base robots, wheeled robots, quadrupted robots, human robots, soft robots, etc. Their unique shapes and designs make them specialized in specific tasks. For example, the fixed-base robots, like franka robots~ \cite{haddadin2022franka} are usually employed in industrial environments for automated picking and placing tasks. In contrast, humanoid robots, with their human-like appearance and adaptability, are versatile and can be deployed across a broad range of domains. 

An autonomous vehicles can also be considered as an embodied agents. They perceive their environment, make real-time decisions, and interact with both drivers and the surrounding environment. Beyond driving safety, these systems are increasingly capable of interpreting human instructions and engaging in conversations with passengers \cite{cui2024survey}. Virtual agents, on the other hand, are prominent in applications such as gaming \cite{hubble2021artificial}, social experiments \cite{park2023generative}, virtual idols \cite{kong2021difference}, etc. These agents interact with users through multiple modalities, including language, visual, and audio, enabling rich and immersive experiences. 

\subsection{Large Language Models}
LLMs, such as GPT-4 \cite{GPT4}, BERT \cite{devlin2018bert}, and T5 \cite{2020t5}, have emerged as foundational components in modern AI, particularly within NLP. These models are designed to capture complex linguistic patterns and structures through unsupervised learning on massive amounts of text data. In the context of embodied multimodal systems, LLMs play a critical role as the primary mechanism for understanding and generating natural language. Their ability to process and produce coherent text enables them to bridge the gap between different modalities, including vision, speech, and action. By integrating LLMs with visual or auditory data, embodied systems can interpret multimodal inputs and generate contextually relevant responses, thus enabling more interactive and intelligent behaviors. Essentially, LLMs serve as the ``language brain'' of these systems, enabling them to understand and execute language-based commands, describe visual scenes, or facilitate complex reasoning across modalities.

BERT \cite{devlin2018bert}, proposed by Google, is based on the Transformer architecture and utilizes a masked language model for pre-training, which significantly improves performance across various NLP tasks. Generative Pre-trained Transformer (GPT) \cite{GPT1}, developed by OpenAI, is a Transformer-based generative model that employs autoregressive training to generate text sequentially. GPT employs a large-scale unsupervised learning approach, leading to breakthroughs in generative models. GPT-2 further scales up both the size and performance of language models, showcasing its ability to generate coherent and natural language. XLNet \cite{yang2019xlnet} introduces a model that combines autoregressive and autoencoding approaches, surpassing BERT on multiple NLP benchmarks. Text-to-Text Transfer Transformer (T5) \cite{2020t5}, proposed by Google, unifies all NLP tasks into a ``text-to-text'' framework, enabling the model to perform transfer learning across different tasks.

The release of GPT-3 \cite{GPT3} marked a milestone as the largest and most powerful language model at the time, with 175 billion parameters. GPT-3's launch represented a significant breakthrough in model scale, dramatically enhancing its performance in tasks such as text generation, question answering, and translation, while also demonstrating the potential for zero-shot and few-shot learning. ChatGPT (based on GPT-3.5 and subsequent versions) revolutionized conversational AI, enabling natural and fluid interactions with users while supporting a wide range of knowledge domains. This development marked a turning point in the practical application of LLMs, drawing significant attention from industries such as customer service, education, and content creation. GPT-4 \cite{GPT4} further advanced reasoning capabilities and introduced multi-modal abilities, demonstrating stronger intelligent features through the combined use of text and images. It extended the capabilities of traditional LLMs by supporting multimodal inputs, such as images alongside text. This allowed GPT-4 to generate text descriptions for images, answer questions related to visual content, and even guide actions in embodied systems like robots. DeepSeek-V3 \cite{liu2024deepseek} further expands the boundaries of multimodal reasoning by adopting a dynamic sparse activation architecture. It innovatively introduces a hybrid routing mechanism that combines task-specific experts with dynamic parameter allocation, achieving higher computational efficiency in cross-modal fusion tasks. These models showcase the power of LLMs in creating robust multimodal systems capable of not only understanding but also interacting with the world through both language and sensory inputs. Meta's LLaMA \cite{llama1} series (including LLaMA-2 \cite{llama2}) has also emerged as a competitor, reflecting growing investments by tech companies in the development of LLMs.

The evolution of LLMs has transitioned from simple statistical models to deep learning-based breakthroughs, and now to the ultra-large-scale Transformer models. With the continued enhancement of computational power and algorithm optimization, LLMs are expected to achieve greater levels of intelligence, enabling them to play an increasingly pivotal role across diverse domains. In the domain of embodied intelligence, these models will further enhance the interaction capabilities between agents and humans, making embodied agents smarter and more capable.

\subsection{Large Vision Models}
Unlike LLMs, LVMs process image or video information. These models have demonstrated exceptional performance in tasks such as image recognition, object detection, image generation, and cross-modal learning. In the realm of embodied intelligence, LVMs also play a crucial role, enabling robots to perceive and understand the visual world in complex and dynamic environments.

ResNet \cite{he2016deep} is a deep convolutional neural network, most notably distinguished by the introduction of residual connections. These connections help address the gradient vanishing and gradient exploding issues that can arise when training very deep neural networks, enabling effective training even with networks containing hundreds or thousands of layers.

Vision Transformer (ViT) \cite{dosovitskiy2020image} is a groundbreaking model that applies the Transformer architecture, originally developed for NLP, to computer vision. Instead of relying on traditional convolutional operations, ViT divides images into fixed-size patches and processes these patches using a self-attention mechanism to capture global dependencies across the image. ViT demonstrates that the Transformer architecture is not only effective for NLP but also powerful in computer vision, particularly when working with large-scale datasets. The success of ViT marks the beginning of a shift away from the dominance of convolutional neural networks, with Transformers emerging as a key tool for visual tasks.

Swin Transformer \cite{liu2021swin} is another Transformer-based model in computer vision. Its key innovation is the introduction of a ``window'' mechanism for performing self-attention within local regions, which improves computational efficiency. Unlike ViT, Swin Transformer adopts a hierarchical design that gradually expands the receptive field, along with window-based division, enabling the model to capture local information while preserving global context. Swin Transformer has delivered strong performance across multiple visual tasks, particularly in object detection and semantic segmentation.

Segment Anything Model (SAM) \cite{kirillov2023segment} is a new visual model launched by Meta that aims to provide a general, flexible, and efficient solution for image segmentation tasks. SAM is trained on a large-scale dataset and can handle a variety of segmentation tasks, including semantic segmentation, instance segmentation, and object segmentation. In addition to supporting traditional segmentation tasks, SAM can be fine-tuned based on user interactions, offering strong adaptability. The release of SAM has established a powerful benchmark for segmentation and spurred further research into interactive and adaptive learning models in computer vision.

DINO \cite{caron2021emerging} and DINOv2 \cite{oquab2023dinov2} are self-supervised learning models proposed by the Facebook AI research team, built on the ViT architecture. The key innovation of DINOv2 lies in its self-supervised approach, which enables effective image representation learning without requiring manual labels. Unlike traditional self-supervised methods, DINOv2 employs an enhanced contrastive learning technique and leverages a larger training dataset, improving the model's representational capacity. DINOv2 has achieved top-tier performance across various visual tasks, including image retrieval, classification, and detection, and has opened new avenues for self-supervised learning research in computer vision.

\subsection{Large Vision-Language Models}
Among EMLMs, Large Vision-Language Models (LVLMs) enhance an agent's environmental understanding, reasoning, and task execution by integrating visual and linguistic information. LVLMs enable agents to fuse and coordinate multimodal data, allowing them to recognize objects through visual input and perform actions based on language instructions. Additionally, LVLMs facilitate cross-modal reasoning and adaptive decision-making in dynamic environments, significantly improving the interaction and navigation capabilities of robots.

CLIP \cite{radford2021learning} is a multi-modal model developed by OpenAI, designed to embed both images and text into a shared vector space using contrastive learning techniques. The model is pre-trained on a large corpus of image-text pairs, enabling it to perform tasks such as image classification and image-text retrieval. CLIP employs contrastive loss to optimize the alignment between text and image descriptions, demonstrating strong zero-shot learning capabilities.

DALL·E \cite{ramesh2021zero} is an image generation model released by OpenAI that generates images based on textual prompts. Built upon the GPT-3 and Vector Quantized Variational Autoencoder architectures, the model excels in creating highly realistic and creative images. DALL·E2 \cite{ramesh2022hierarchical} improves upon its predecessor by enhancing the quality and diversity of generated images, enabling the creation of detailed and complex visuals based on sophisticated text descriptions.DALL·E3 \cite{betker2023improving} has better performance.

BLIP \cite{li2022blip}, introduced by Salesforce Research, employs a two-way self-supervised learning approach to integrate visual and linguistic information. BLIP boosts pre-training efficiency through a “guided” strategy, helping the model better grasp the finer details in visual-language tasks, particularly in visual question answering (VQA) and image captioning.

Flamingo \cite{alayrac2022flamingo} is a novel visual-language model from DeepMind that can process multi-modal data (images and text) and perform cross-modal reasoning. Unlike traditional models, Flamingo excels in few-shot learning, allowing it to tackle tasks with minimal labeled data without the need for extensive training datasets. This model emphasizes few-shot capabilities, making it suitable for scenarios that require efficient handling of various data types.

Visual BERT \cite{li2019visualbert} is a variant of the BERT model developed by Facebook AI Research that integrates visual information. It jointly encodes image features and text to address cross-modal tasks. By embedding image region features into the BERT framework, Visual BERT optimizes the semantic alignment between text and images during pre-training.

ALIGN \cite{jia2021scaling} is a large-scale image-text alignment model from Google Research. It leverages contrastive learning on vast amounts of image-text data from the internet to understand the relationships between images and their textual descriptions. ALIGN excels in tasks like image classification and image-text retrieval, offering powerful cross-modal search capabilities through large-scale training.

GIT \cite{wang2022git} is a pre-trained model from Microsoft Research designed for image-text generation and understanding. It efficiently extracts and generates information across both visual and linguistic domains. GIT can generate descriptive text from images and create corresponding images based on text inputs, facilitating versatile cross-modal generation tasks.

MDETR \cite{kamath2021mdetr} is a Transformer-based visual-language model developed by Facebook AI Research that uses a ``modulation mechanism'' to handle image-text pair tasks. The model identifies specific regions in images based on text descriptions, allowing for precise multi-modal reasoning by associating these regions with the provided text.

PaLM-E \cite{driess2023palm} is a powerful multimodal pre-training model proposed by Google Research that combines three modalities: images, text, and actions (such as robotic operations) to promote the development of cross-modal intelligent systems. PaLM-E is an extended version of PaLM (Pathways Language Model), focusing on bridging the gap between vision and language models while providing enhanced perception and reasoning capabilities for robots and other intelligent systems.

CoCa \cite{yu2022coca} was released by Meta. One of its important features is that it combines contrastive learning and generative learning. Contrastive learning is usually used to improve the performance of models in visual-language tasks, helping models to better understand and distinguish the relationship between images and text. Generative learning enables the model to generate descriptive text based on the input image, or generate images based on text.

\subsection{Other Modal Models}
\subsubsection{Vision-Audio Models}

In addition to vision and language, audio plays a crucial role in everyday tasks, helping us identify scenarios and locate sound-emitting objects. While most research focuses on vision-audio or audio-language data, few studies explore audio in embodied tasks. Recent work in audio-visual navigation tasks, such as SoundSpaces \cite{chen2020soundspaces}, combines vision and audio for tasks like AudioGoal (where targets are indicated by sound) and AudioPointGoal (where audio provides directional guidance). These tasks face challenges, such as the need for continuous sound, which has been addressed by linking audio with semantic meanings and spatial properties \cite{chen2021semantic}. Researchers have also introduced more complex audio scenarios with multiple sound sources and distractions \cite{yu2022sound,younes2023catch}. In manipulation tasks, audio provides crucial contact information, complementing vision and touch \cite{li2023see}. Studies show that combining audio with action generation and feedback significantly improves task performance, with self-supervised learning methods and augmented audio data boosting success rates in real-world environments~ \cite{thankaraj2023sounds,liu2024maniwav}.

\subsubsection{Vision-Touch Models}

In manipulation tasks, visual information is often the primary source for adjusting robot motions, but vision sensors can be limited by occlusion and their inability to measure contact force, which is crucial for successful actions. To address this, several studies have explored combining vision and tactile data. For example, \cite{dikhale2022visuotactile} proposed a network that integrates vision and tactile data to estimate the 6D pose of objects for in-hand manipulation. In grasping tasks, \cite{calandra2018more} investigated a learned regrasp policy that iteratively adjusts the grasp using both visual and tactile feedback, showing improved performance with touch information. To handle more complex objects, such as deformable ones, \cite{han2024learning} introduced a Transformer-based framework that uses tactile and visual data for safe grasping, employing exploratory actions to gather tactile feedback and predict the grasp outcome for safer parameter selection.

\section{Development of Embodied Multimodal Large Models}
\label{sec:EMLM}
\begin{figure}[t]
	\centering
	\includegraphics[width=1.0\linewidth, trim={0
	0 90 20}, clip]{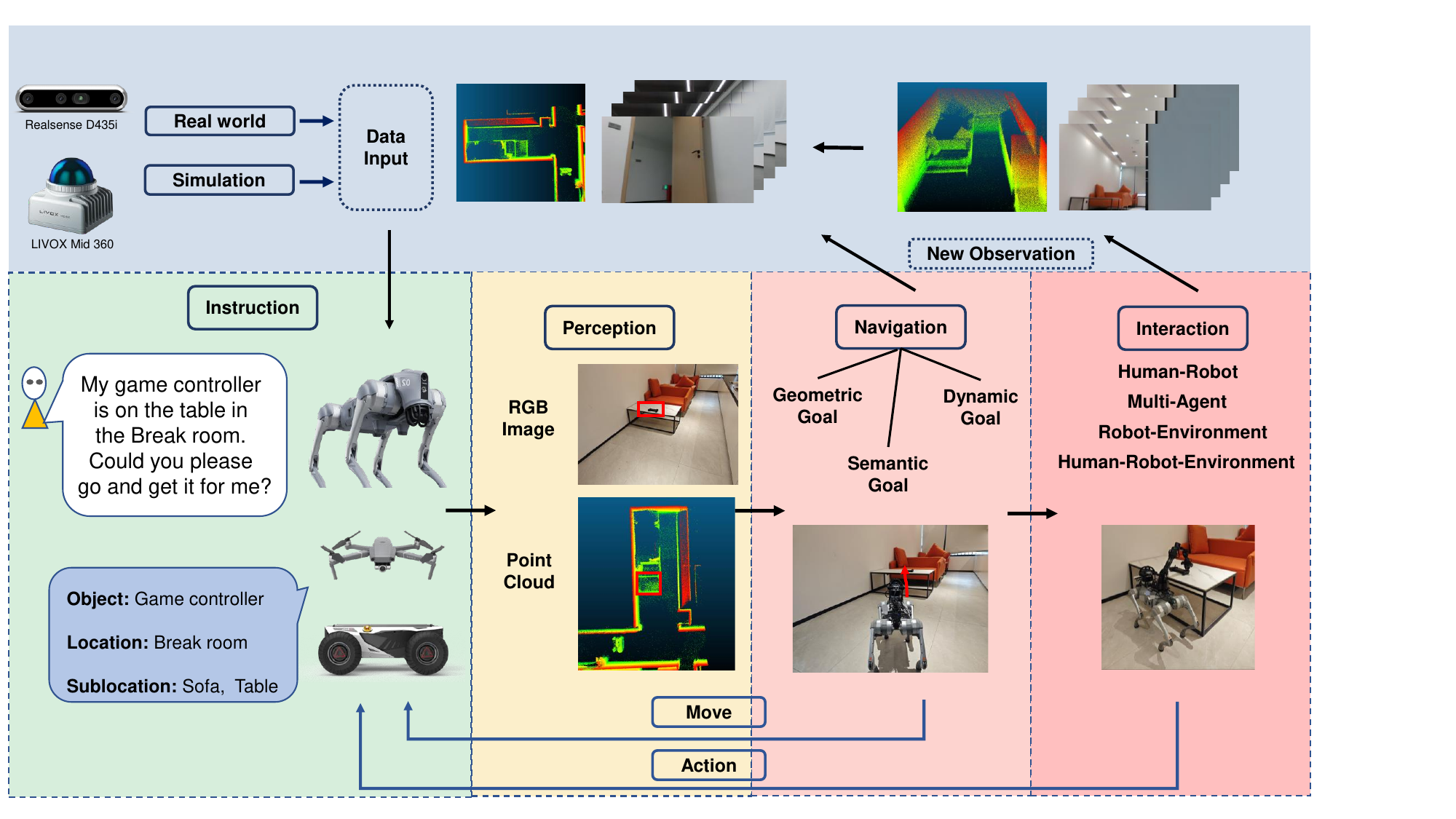}
	
	\caption{Full Task Stacks for Embodied Agents. Various embodied intelligent agents, including robot dogs, humanoid robots, and other types of intelligent systems, rely on a range of sensors, such as cameras, LiDAR, and other sensing technologies, to perceive their environment. These agents then perform specific tasks, usually guided by human voice or language commands. Task execution typically involves three key modules: perception, navigation, and interaction. The datasets and large models required for these modules can be collected and trained using either simulators or real-world scenarios. During task execution, the agent interacts with its environment to gather the necessary information.}
	\label{fig:embodyRobotTask}
\end{figure}

EMLMs are a type of AI models that combine multiple modal information such as language, vision, and hearing. They can understand and process different types of data from the real world. As shown in Fig. \ref{fig:embodyRobotTask}, these models are usually designed to perform various tasks, such as perception, navigation and interaction, etc.

\subsection{Embodied Perception}
Different from using traditional neural network methods or large models to identify objects, according to the definition of embodied intelligence, embodied agents have the ability to interact with and move in the physical world. This requires EMLMs to have a deeper perception and understanding of objects in the physical world and the motion and logical relationships between objects. Embodied perception requires visual perception and reasoning, understanding 3D relationships in the scene, predicting and executing complex tasks based on visual information. The development of the embodied perception large model is shown in Fig. \ref{fig:Section3}

Currently, there are two main types of large models for embodied perception: one is based directly on GPT, and the other is based on other large models. The detailed information of these models can be found in Table \ref{tab:Perception}.

\begin{table}[t]
    \centering
    \caption{Embodied perception large models}
    \label{tab:Perception}
    \resizebox{1\columnwidth}{!}{
    \begin{tabular}{c >{\centering\arraybackslash}m{3cm} c >{\centering\arraybackslash}m{4cm} c >{\centering\arraybackslash}m{3cm} c}
        \toprule
        Model & Architecture & Size & Platform & Dataset & Hardware & Year \\
        \toprule
            Octopus \cite{yang2025octopus} & MPT-7B, CLIP VIT-L/14 \cite{radford2021learning} & - & (Sim) OctoGibson \cite{li2023behavior}, OctoGTA & OctoGibson, OctoGTA & - & 2025 \\
            \addlinespace[0.5em]
            AlphaBlock \cite{jin2023alphablock} & Vicuna \cite{chiang2023vicuna}, ViT-G/14 \cite{fang2023eva} & - & Franka Emika Research 3 robot arm & AlphaBlock & 2 RTX A6000 & 2023 \\
            \addlinespace[0.5em]
            CoPa \cite{huang2024copa} & GraspNet \cite{fang2020graspnet}, GPT-4V & - & Franka Emika Panda & - & - & 2024 \\
            \addlinespace[0.5em]
            ReKep \cite{huang2024rekep} & DINOv2 \cite{oquab2023dinov2}, GPT-4o \cite{islam2024gpt} & - & Wheeled Single-Arm Platform and Stationary Dual-Arm Platform & - & - & 2024 \\
             \midrule
            RobotGPT \cite{jin2024robotgpt} & - & - & PyBullet, MoveIt and ros franka & - & - & 2024 \\
            \addlinespace[0.5em]
            Voxposer \cite{huang2023voxposer} & - & - & Franka Emika Panda & real, SAPIEN \cite{xiang2020sapien} & - & 2023 \\
            \addlinespace[0.5em]
            PaLM-E \cite{driess2023palm} & ViT, OSRT \cite{sajjadi2022object} & - & TAMP, Language-Table, Mobile Manipulation & - & - & 2023 \\
             \midrule
            RT-1 \cite{brohan2022rt} & EfficientNet-B3 \cite{tan2019efficientnet} & - & Everyday Robots & RT-1 & - & 2022 \\
            \addlinespace[0.5em]
            RT-2 \cite{brohan2023rt} & PaLI-X, PaLM-E & - & 7DoF mobile manipulator & - & - & 2023 \\
            \addlinespace[0.5em]
            RT-H \cite{belkhale2024rt} & ViT \cite{dosovitskiy2020image}, PaLI-X \cite{chen2023pali} & 55B & - & - & - & 2024 \\
            \addlinespace[0.5em]
            RoboFlamingo \cite{li2023vision} & OpenFlamingo \cite{awadalla2023openflamingo} & - & Franka Emika Panda & CALVIN \cite{mees2022calvin} & 8 NVIDIA Tesla A100 & 2023 \\
            \midrule
            Openvla \cite{kim2024openvla} & Prismatic-7B \cite{karamcheti2024prismatic} & 7B & WidowX robot & BridgeData V2 \cite{walke2023bridgedata} & 21,500 A100-hours and RTX 4090 (6HZ) & 2024 \\
            RoboMamba \cite{liu2024robomamba} & Mamba \cite{gu2023mamba} & - & Franka Emika Panda, SAPIEN & \begin{tabular}[c]{@{}c@{}}LLaVA-LCS \cite{liu2024improved},\\ LLaVA-v1.5 \cite{liu2024improved},\\ LRV-INSTRUCT \cite{liu2023mitigating},\\ RoboVQA \cite{sermanet2024robovqa}\end{tabular} & NVIDIA A100 & 2024 \\
        \bottomrule
        \end{tabular}
    }
\end{table}

\begin{figure}[t]
	\centering
	\includegraphics[width=1.0\linewidth]{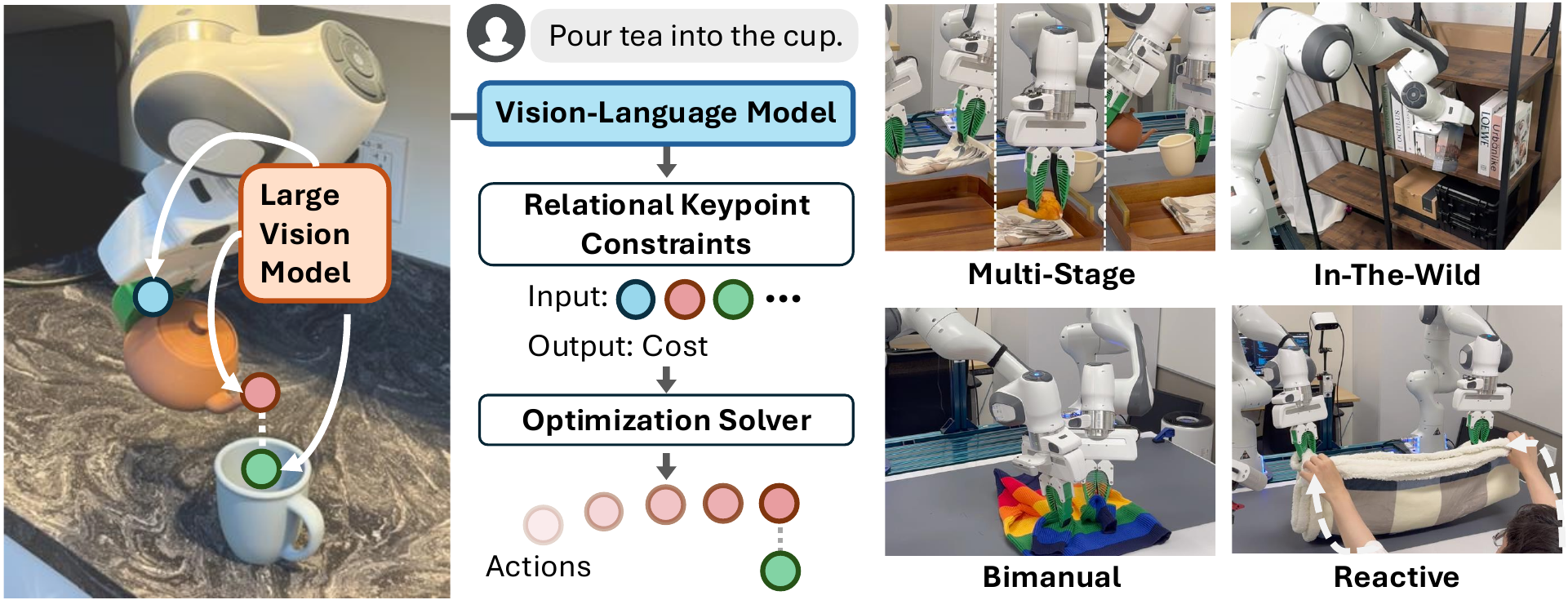}
	
	\caption{Diagram illustrating the Rekep framework (Source: Rekep \cite{huang2024rekep}).}
	\label{fig:rekep}
\end{figure}

\subsubsection{GPT-Based Large Models}
The general large model can accept specific text instructions tailored to the task's requirements, returning scene understanding results in a natural language format and handling a variety of perception tasks. For example, models like GPT1 \cite{GPT1}, GPT2 \cite{GPT2}, GPT3 \cite{GPT3}, GPT4 \cite{GPT4} can effectively perform these functions.

Octopus \cite{yang2025octopus} uses GPT-4V \cite{yang2023dawn} to dynamically generate descriptions and analyses of observed images based on the current stage task, including objects that can be interacted with in the scene and their relative positions in space. This description is used as input to the language model to generate decisions for the next action. Also using GPT-4V, CoPa \cite{huang2024copa} utilizes GPT-4V to directly identify and highlight the specific grasping areas and potential grasping poses of generated objects in the observed image, enabling fine-grained object understanding. Subsequently, GraspNet \cite{asif2018graspnet} acts as a grasping pose detector, selecting the pose with the highest confidence for execution.

However, researchers have observed that GPT-4V frequently struggles to generate satisfactory results when provided with coordinate-based inputs. This limitation is attributed to the inherent constraints of the text-based GPT-4 model, which lacks robust spatial perception capabilities. Then, they propose AlphaBlock \cite{jin2023alphablock}, which was inspired by method chain-of-thought \cite{wei2022chain}. This approach requires GPT-4V to engage in reasoning before generating coordinates, ensuring it also provides an understanding and description of the layout. This perceptual information is then used to deliver real-time feedback on the execution status of the current task. To ensure accurate perception of patch features (e.g., color and spatial location), they designed a visual adapter to extract and merge multi-stage visual features from the ViT \cite{dosovitskiy2020image} model in MiniGPT-4 \cite{zhu2023minigpt}. To ensure consistent embeddings in LLMs, they further adopted a visual Q-former \cite{li2023blip} with a language-specific projector to effectively align image observations with LLMs.

Some methods directly use the observation and code generation capabilities of multimodal large models to output executable code by observing input. For example, Voxposer \cite{huang2023voxposer} employs a multimodal large model to extract the spatial geometric information of an object and generate its 3D coordinates. These object coordinates are then used to populate parameters in the code, facilitating the generation of a series of 3D functional diagrams and constraint diagrams based on the robot's observational space.

Compared with Voxposer's method, which uses direct visual observation, Rekep \cite{huang2024rekep} is more adaptable to environmental changes and has a higher degree of automation. ReKep offers significant improvements. ReKep \cite{huang2024rekep} uses DINOv2 \cite{oquab2023dinov2} to extract patch features of the input RGB image and samples it to the original image size through bilinear interpolation. Then, Segment Anything Model \cite{kirillov2023segment} is used to extract all the markers in the scene and obtain candidate key points. After obtaining the key point candidates, researchers superimpose them on the original RGB image together with the numerical markers. Next, the image and task instructions are fed into GPT-4o \cite{islam2024gpt}, where a specific prompt is employed to generate the required number of stages, along with the corresponding sub-goals and path constraints for each stage.

Similarly, RobotGPT \cite{jin2024robotgpt} also uses GPT to directly generate executable code. It proposes a five-part prompting approach, consisting of background information, object details, environmental context, task specifications, and examples. By providing detailed descriptions of the environment, objects, and tasks, this method effectively guides ChatGPT to generate precise and relevant outputs. The generated code is then executed to control the robot in performing the specified tasks.

\subsubsection{Method Based on Non-GPT Large Models}
PaLM-E \cite{driess2023palm}, which directly integrates continuous inputs from the sensor modalities of an embodied agent, allowing the language model to make more grounded inferences for sequential decision-making in the real world. Inputs such as images and state estimates are embedded into the same latent space as language tokens, enabling them to be processed by the self-attention layers of a Transformer-based \cite{vaswani2017attention} LLMs in the same manner as text. ViT \cite{dosovitskiy2020image}

RT-1 \cite{brohan2022rt} employs the EfficientNet-B3 \cite{tan2019efficientnet} model as a perception module to encode image data. It transforms images into implicit representations for subsequent task execution. Through training, this perception module learns to encode information in a way that enhances task execution effectiveness. As an upgrade of RT-1, RT-2 \cite{brohan2023rt} introduces a method that directly trains a vision-language model to output low-level robot actions by representing these actions as textual tokens. This approach involves training the model alongside Internet-scale vision-language tasks to improve its performance and versatility. After combining the advantages of RT-1 and RT-2 and improving them, RT-H \cite{belkhale2024rt} instantiates this LVLM using the same PaLI-X 55B \cite{chen2023pali} architecture as RT-2. It first tags the image with the ViT \cite{dosovitskiy2020image} encoder model, and then employs an encoder-decoder Transformer to convert the image and natural language tag streams into action tags.

RoboFlamingo \cite{li2023vision} is built on the open-source vision-language model, OpenFlamingo \cite{awadalla2023openflamingo}, and is tailored for downstream manipulation tasks through a small amount of imitation learning fine-tuning using robot manipulation data. Its vision encoder consists of ViT \cite{dosovitskiy2020image} and Perceiver Resampler \cite{alayrac2022flamingo}. Its flexible design allows it to be deployed on lower-performance platforms. 

Openvla \cite{kim2024openvla} is a 7B-parameter open-source Vision-Language-Action (VLA) trained on a diverse dataset of 970,000 real-world robot demonstrations. It enhances the Llama 2 \cite{touvron2023llama} language model by incorporating a visual encoder that integrates pretrained features from DINOv2 \cite{oquab2023dinov2} and SigLIP-grade \cite{zhai2023sigmoid} GPUs. Among them, DINOv2 focuses on spatial reasoning to provide fine-grained visual information, while SigLIP focuses on semantic understanding to enrich visual features.

RoboMamba \cite{liu2024robomamba} integrates a visual encoder with the Mamba model \cite{gu2023mamba}, aligning visual data with language embeddings through co-training. This enables the model to develop visual common sense and reasoning capabilities related to robotics.


\subsection{Embodied Navigation}
Relative to traditional A-to-B robot navigation, where an agent is instructed to move from Room A (x1, y1) to Room B (x2, y2), and the agent uses A* \cite{hart1968formal} or Dijkstra's \cite{dijkstra2022note} algorithm on a pre-generated 2D map to find a reachable shortest path, embodied intelligent navigation systems adopt a more dynamic and environmentally perceptive approach. These navigation systems do not solely rely on static map data but perceive and process surrounding environments in real-time through sensors, and models, converting environmental information into understandable and actionable semantics for the agent.

In the realm of navigation with large-scale embodied intelligence models, two primary methodologies are prevalent: the first approach is to harness a general large model, whereas the second entails crafting a specialized, EMLMs tailored explicitly for embodied intelligence tasks.

\begin{table}[H]
    \centering
    \caption{Embodied navigation Large models}
    \label{tab:navigation}
    \resizebox{1\columnwidth}{!}{
    \begin{tabular}{c >{\centering\arraybackslash}m{3cm} c >{\centering\arraybackslash}m{4cm} >{\centering\arraybackslash}m{3cm} >{\centering\arraybackslash}m{3cm} c}
        \toprule
        Model & Architecture & Size & Platform & Dataset & Hardware & Year \\
        \toprule
            DiscussNav \cite{DiscussNav} & GPT-4V \cite{yang2023dawn}, InstructBLIP \cite{InstructBLIP} & unpublished/13B & Turtlebot4Lite &  R2R \cite{anderson2018vision}  &  -  &  2024  \\
            Trans-EQA \cite{DiscussNav} & GPT-4V \cite{yang2023dawn}, InstructBLIP \cite{InstructBLIP} & unpublished/13B & Turtlebot4Lite &  R2R \cite{anderson2018vision}  &  -  &  2024  \\
            LM-Nav \cite{shah2023lm} &  GPT-3 \cite{GPT3} & 175B & Clearpath Jackal UGV & Real-world environments & - &  2023\\
            \midrule
            L3MVN \cite{yu2023l3mvn} &  RoBERTa-large \cite{liu2019roberta},CLIP \cite{radford2021learning} & 3.55B/191M & Jackal Robot & HM3D \cite{ramakrishnan2021habitat}, Gibson \cite{xia2018gibson} & - &  2023\\
            \midrule
            LFG \cite{shah2023navigation} &  GPT-3.5 \cite{ouyang2022training} & 175B & LoCoBot & HM3D \cite{ramakrishnan2021habitat} & 4 NVIDIA V100 GPUs &  2023\\
            
            VLMaps \cite{huang2023visual} &  LSeg \cite{li2022language}, GPT-4V \cite{yang2023dawn} & 307M/unpublished & HSR mobile & MP3D \cite{chang2017matterport3d}, AI2THOR \cite{kolve2017ai2} & - &  2023\\
            NavGPT \cite{zhou2024navgpt} &  GPT-3.5 \cite{ouyang2022training},BLIP-2 \cite{li2023blip} & unpublished/188M & simulate & R2R \cite{anderson2018vision} & - &  2024\\
            \addlinespace[0.5em]
            SG-Nav \cite{yin2024sg} &  LLaMA \cite{llama1},GPT-4 \cite{GPT4}  & 7B/unpublished & simulate & MP3D \cite{chang2017matterport3d}, HM3D \cite{ramakrishnan2021habitat}, RoboTHOR \cite{deitke2020robothor} & - &  2024\\
            \midrule
            Simple but effective \cite{khandelwal2022simple} &  CLIP \cite{radford2021learning}  & 191M & simulate & RoboTHOR \cite{deitke2020robothor}, HP3D \cite{ramakrishnan2021habitat}, iTHOR \cite{weihs2021visual}  & - &  2020\\
            \addlinespace[0.5em]
            Zson \cite{majumdar2022zson} &  CLIP \cite{radford2021learning}  & 191M & simulate & Gibson \cite{xia2018gibson}, HM3D \cite{ramakrishnan2021habitat}, MP3D \cite{chang2017matterport3d} & 8 NVIDIA A40 GPUs &  2022\\
            \midrule
            NavGPT-2 \cite{zhou2025navgpt} &  InstructBLIP \cite{InstructBLIP}  & 5B & simulate & R2R \cite{anderson2018vision}, RxR \cite{ku2020room}, HM3D \cite{ramakrishnan2021habitat} & NVIDIA A100 GPU &  2025\\
            
            vlfm \cite{yokoyama2024vlfm} &  BLIP-2 \cite{li2023blip}  &  188M  &  simulate  & Gibson \cite{xia2018gibson}, HM3D \cite{ramakrishnan2021habitat}, MP3D \cite{chang2017matterport3d}  &  RTX 4090 MaxQ Mobile GPU  &  2024\\
            \addlinespace[0.5em]
            NavCoT \cite{lin2024navcot} &  LLaMA \cite{llama1},BLIP  & 7B & simulate & R2R \cite{anderson2018vision}, RxR \cite{ku2020room}, R4R \cite{jain2019stay}, REVERIE \cite{Qi_2020_CVPR} & 4 NVIDIA V100 GPUs &  2024\\
            \midrule
            NaviLLM \cite{zheng2024towards} &  Vicuna-7B-v0 \cite{zheng2023judging},ViT \cite{dosovitskiy2020image}  & 7B/428M & simulate & \parbox{3cm}{\centering CVDN \cite{thomason2020vision}, SOON \cite{zhu2021soon}, R2R \cite{anderson2018vision},\\REVERIE \cite{Qi_2020_CVPR}, ScanQA \cite{Azuma_2022_CVPR},\\LLaVA-23k \cite{liu2024visual}, MP3D-EQA \cite{wijmans2019embodied}} & 8 NVIDIA V100 GPUs &  2024\\
            \midrule
            GOAT \cite{wang2024vision} &  GOAT  & - & simulate & R2R \cite{anderson2018vision}, RxR \cite{ku2020room}, REVERIE \cite{Qi_2020_CVPR}, SOON \cite{zhu2021soon} & NVIDIA V100 GPU &  2024\\
            \addlinespace[0.5em]
            VER \cite{liu2024volumetric} &  ViT-B/16 \cite{dosovitskiy2020image}  & 86M & simulate & MP3D \cite{chang2017matterport3d}, R2R \cite{anderson2018vision}, REVERIE \cite{Qi_2020_CVPR}, R4R \cite{jain2019stay} & 8 NVIDIA RTX 4090 GPUs & 2024\\
            \addlinespace[0.5em]
            GNM \cite{shah2022gnm} &  MobileNetv2 \cite{sandler2018mobilenetv2}  & 6.9M & Vizbot,DJI Tello,Clearpath Jackal UGV,LoCoBot. & Real-world environments & - &  2023\\
            \addlinespace[0.5em]
            ViNT \cite{shah2023vint} &  ViNT  & 31M & Vizbot,Unitree Go 1,Clearpath Jackal UGV,LoCoBot. & Real-world environments & 8 NVIDIA V100 GPUs & 2023\\
            \addlinespace[0.5em]
            NoMaD \cite{sridhar2023nomad} &  NoMaD  & 19M & LoCoBot & Real-world environments & - & 2023\\
        \bottomrule
        \end{tabular}
    }
\end{table}

\begin{figure}[t]
	\centering
	\includegraphics[width=1.0\linewidth]{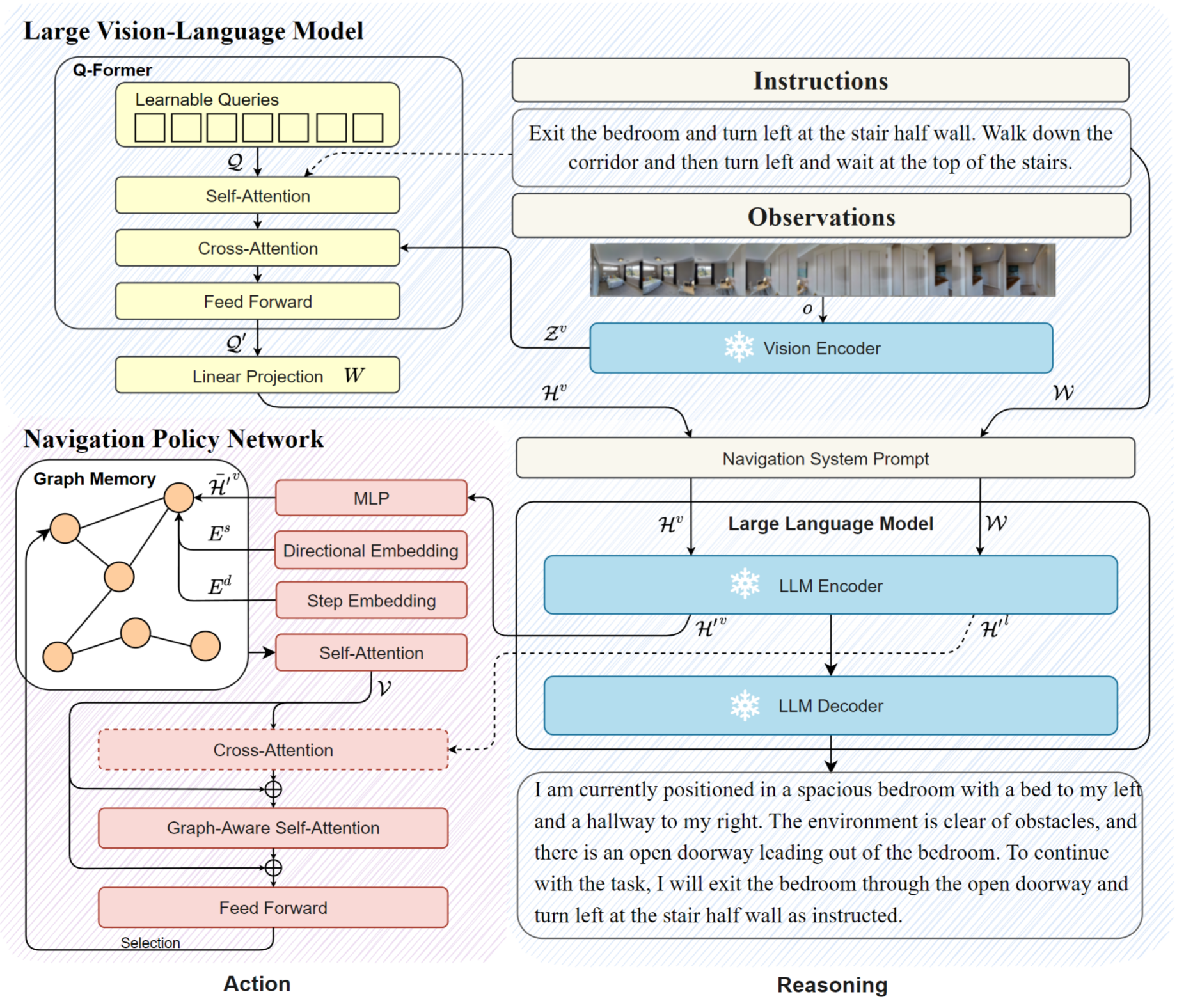}
	
	\caption{Diagram illustrating the NavGPT-2 framework (Source: NavGPT-2 \cite{zhou2025navgpt}).}
	\label{fig:NavGpt-2}
\end{figure}

\subsubsection{General Large Models}
LVLMs are capable of understanding and generating natural language due to their vast scale, advanced architecture, and pre-training on massive datasets. In some cases, they demonstrate reasoning and semantic understanding abilities that approach or even surpass those of humans. The excellent reasoning ability of LLMs can transform abstract instructions into operable semantic topologies. For example, LM-Nav \cite{shah2023lm} implements natural language instruction parsing based on GPT-3 \cite{GPT3}, making navigation decisions by extracting textual landmarks and combining them with scene images. Its research focuses on the application of LLMs in natural language understanding. In contrast, L3MVN \cite{yu2023l3mvn} demonstrates two innovative paradigms of LVLMs in navigation tasks through the RoBERTa-large model \cite{liu2019roberta}: The zero-shot method constructs semantic description sentences such as ``This scene contains a bathtub, toilet, and TV,'' and uses the LVLMs to evaluate the probability of the navigation target's existence; the feed-forward method encodes query sentences into summary embeddings, which are then input into a neural network for target probability prediction.Through experiments in real-world environments for the former and in different simulated environments (HM3D \cite{ramakrishnan2021habitat}, Gibson \cite{xia2018gibson}) for the latter, the transferability and flexibility of LVLMs in different scenarios and usage methods are demonstrated.

In the paper \cite{shah2023navigation}, as well as in VLMaps \cite{huang2023visual}, a method similar to the LLMs approach used in L3MVN's Zero-shot paradigm was also employed. The former proposed the Language Frontier Guide (LFG), which differs by not only using positive prompt sentences but also incorporating negative prompt sentences, and obtaining the likelihood estimates of the target through both \cite{wei2022chain}. The latter, on the other hand, utilizes GPT-4v \cite{yang2023dawn} to enable the LLMs to speculate which known object in the catalog is most likely to be near the target.

Recent advancements further expand LLMs' architectural roles in embodied navigation through system-level integration and structured environment comprehension. These innovations transcend basic semantic parsing by embedding LLMs into multimodal reasoning pipelines while introducing self-corrective mechanisms for enhanced reliability.

NavGPT \cite{zhou2024navgpt} is a system entirely based on LLMs for understanding and executing natural language navigation instructions. The core of this work lies in its ability to handle multimodal inputs, including text descriptions of visual observations, navigation history, and potential future exploration directions. By leveraging the GPT-3.5 \cite{ouyang2022training} model to summarize historical information, it generates concise prompts for the LLMs to make decisions. This work demonstrates the potential of LLMs in processing multimodal inputs, conducting advanced planning, and providing interpretable reasoning processes. However, for purely visual modality inputs, other Visual Frontend Models, are still required for preprocessing.

The SG-Nav framework \cite{yin2024sg} constructs a hierarchical 3D scene graph and utilizes LLMs for hierarchical chain-of-thought prompting (H-CoT), enabling LLMs to deduce the location of target objects based on the structural information of the scene graph. This process includes predicting the distance between objects and targets, posing relevant questions, answering questions, and predicting the distance between subgraphs and targets based on subgraph information. Additionally, SG-Nav has designed a graph-based re-perception mechanism to assess the credibility of detected target objects, and it abandons the target when the credibility is low to correct perception errors, thereby enhancing the accuracy and robustness of navigation.

Language-Image Pre-training model, is pre-trained using a large number of image-text pairs to learn the correlations between the two, achieving cross-modal semantic understanding. It is highly suitable for processing images obtained by an agent from the environment during navigation tasks, to establish correspondences between images and textual instructions or landmarks, for the execution of more advanced and complex navigation tasks.

CLIP \cite{radford2021learning} has collected 400 million image-text pairs from the internet for model learning, achieving an accuracy of 75.4\% on ImageNet \cite{deng2009imagenet}. CLIPORT \cite{shridhar2022cliport} utilizes CLIP's image encoder to provide semantic understanding for robotic manipulation and to form a semantic stream. The features of the semantic stream are up-sample and undergo an element-wise multiplication operation with the features of the spatial stream, which integrates semantic information into the spatial features. CLIP's image encoder can also be combined with Reinforcement Learning (RL) \cite{majumdar2022zson,khandelwal2022simple}, by converting images into feature vectors rich in semantic information. These feature vectors can serve as inputs for RL algorithms, helping robots better understand their environment and make more informed decisions. Combining CLIP with topological and metric maps used in traditional navigation can achieve effects similar to those of RL and end-to-end models while significantly reducing training time and costs. In L3MVN \cite{yu2023l3mvn}, the embeddings formed by image encoding are embedded in the nodes of the topological map, and then text or language instructions are transformed into embeddings through text encoding. By calculating the cosine similarity, the node where the instruction target is located can be determined. VLMaps \cite{huang2023visual} further embeds these embeddings into 3D space, and by applying a top-down compression, it is possible to obtain semantic or obstacle 2D maps suitable for different types of robots, such as vacuum cleaners and drones.

Compared to CLIP, BLIP and BLIP-2 \cite{li2022blip, li2023blip} offer enhanced capabilities. Specifically, BLIP-2 is able to generate text directly from images, enabling the model to produce corresponding textual descriptions based solely on the content of the input images, without the need for additional training data. In NavGPT \cite{zhou2024navgpt}, by inputting images from multiple perspectives, BLIP-2 can generate a preliminary understanding of the current environment, and then GPT-3.5 makes decisions based on historical information and explorable areas. In subsequent research, a fine-tuned model NavGPT-2 \cite{zhou2025navgpt} was constructed based on InstructBLIP \cite{InstructBLIP} by applying the visual encoder of EVA-CLIP \cite{fang2023eva}, which retained the reasoning capabilities of LLMs. NavGPT-2 has matched the performance of state-of-the-art specially pre-training model for Visual Language Navigation (VLN) during the unseen test phase. Similar to NavGPT, vlfm \cite{yokoyama2024vlfm} utilizes BLIP-2 to obtain the cosine similarity scores between the current RGB observation frame and each object in a list of text prompts containing the target, thereby generating a 2D value map. This map is then combined with the Frontier value Map generated in another step to determine the optimal Waypoint.

\subsubsection{Specialized Embodied Intelligence Large Models}
There is a type of model that does not use LVLMs directly. Instead, it combines different models and uses a dedicated dataset to train a new specialized large model.

NavCoT \cite{lin2024navcot} combines existing VLN datasets with advanced models such as LLMs and CLIP to train the LLaMA \cite{touvron2023llama,zhang2023llama} model on extracting object text from multi-perspective images, thereby enhancing its performance on the Future Imagination task and ultimately improving the entire navigation decision-making process.

NaviLLM \cite{zheng2024towards} extracts features from six perspective images at each location using a ViT \cite{dosovitskiy2020image} to form a scene encoding, then uses a Transformer Encoder to capture the interdependencies between different viewpoints. It processes and generates textual data such as task instructions, observation results, and historical information using an LLM (Vicuna-7B-v0 \cite{zheng2023judging}), and finally trains the model with all three as inputs.

Similarly, Trans-EQA \cite{luo2024transformer} leverages the global modeling advantages of Transformer \cite{vaswani2017attention} in the navigation module to replace the local feature extraction bottleneck of traditional CNNs, effectively associating dispersed visual features with language semantics.

GOAT \cite{wang2024vision} decomposes vision, instruction, and history into mediator, observable confounder, and unobservable confounder. It proposes Back-door Adjustment Causal Learning (BACL) and Front-door Adjustment Causal Learning (FACL) causal learning modules to process the corresponding information, while the mediator is used to predict the output and reduce the impact of confounding factors. This enables the GOAT model to have excellent generalizability and the ability to handle dataset biases and reduce the impact of confounding factors. The GOAT model can provide accurate navigation in diverse and unseen environments, making it highly suitable for application in complex real-world scenarios.

Rui Liu et al. \cite{liu2024volumetric} proposed the use of voxels to achieve a more comprehensive 3D representation. Specifically, they used ViT-B/16 \cite{dosovitskiy2020image} during pre-training to extract features from multi-perspective images, introduced cross-view attention (CVA) for coarse sampling between 2D and 3D, and then employed 3D deconvolution to enhance the resolution, realizing a transition from coarse to fine-grained 3D representation. During the training phase of navigation tasks, they froze the weights of some pre-trained layers. The purpose of this approach is to retain the general features learned by the model during the pre-training phase while allowing the model to adapt to specific navigation tasks by adjusting the weights of other layers.

The General Navigation Models introduced by the Berkeley AI Research team: GNM \cite{shah2022gnm},  Visual Navigation Transformer(ViNT) \cite{shah2023vint}, and Navigation Mask Diffusion Strategy(NoMaD) \cite{sridhar2023nomad}, are a set of generalizable models capable of controlling a variety of different robots without the need for specific prior training. GNM achieves broad generalization across multiple robotic platforms by training on a 60-hour heterogeneous dataset collected from various different yet structurally similar robots, including indoor and outdoor environment navigation on unseen robots. ViNT is a Transformer-based \cite{vaswani2017attention} visual navigation foundational model that learns general navigation capabilities through pre-training and can adapt to various downstream tasks, including zero-shot deployment and new task adaptation, demonstrating transfer learning capabilities across diverse datasets. The model predicts the number of time steps (dynamic distance) and a sequence of actions required to reach the goal by encoding current and past visual observations and goals, which are consistent across different robotic platforms, thus enabling rapid and efficient inference for resource-constrained robots as well as the ability to prompt and fine-tune for downstream tasks. NoMaD is a novel robotic navigation model that integrates a Transformer backbone network \cite{vaswani2017attention} with diffusion models \cite{ho2020denoising}, capable of unified processing of goal-oriented navigation and task-agnostic exploration, enabling robots to both execute specified tasks and conduct effective exploration in unknown environments. Through experiments on real-world mobile robot platforms, NoMaD has demonstrated superior performance and lower collision rates compared to existing methods, while also featuring a smaller model size and higher computational efficiency.

\subsection{Embodied Interaction}

Traditional robot interaction methods typically require the integration of independent modules such as perception, decision-making, planning, and control to accomplish specific tasks. With the advancements in deep learning, particularly the significant progress of language and visual models, embodied intelligent interaction has become feasible. Embodied intelligent interaction involves enabling intelligent agents and large models to possess multi-modal processing capabilities, including natural language reasoning, visual-spatial semantic perception, and the alignment of visual perception with language systems, among other key technologies. Currently, the foundational capabilities of embodied intelligent interaction require that the system understand human natural language instructions and autonomously complete tasks. As such, language-based embodied intelligent interaction has become a central focus of research. This can be broadly categorized into two types: language-based short-horizon action policies and language-based long-horizon action policies. Some interaction models are shown in Table \ref{tab:Interaction} and Table \ref{tab:Perception}.

\subsubsection{Language-based Short-horizon Action Policy}

\begin{figure}[t]
	\centering
	\includegraphics[width=1.0\linewidth]{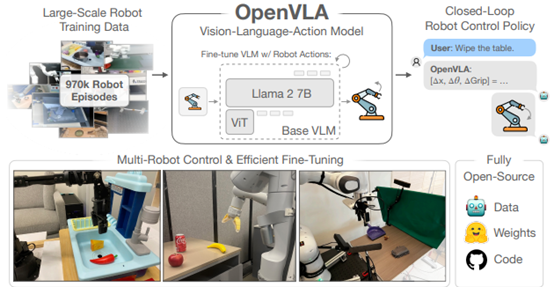}
	
	\caption{Diagram illustrating the Openvla framework (Source: Openvla \cite{kim2024openvla}).}
	\label{fig:Openvla}
\end{figure}

Currently, many researchers focus on language-based vision fusion action strategies for simple tasks; while some researchers improve visual encoders to make them more suitable for robot operations. In the context of embodied interaction, the agent completes the task by interacting with the environment. Short-horizon action strategies emphasize making quick action decisions in a shorter time scale, usually based on the current state and immediate feedback. This strategy does not consider long-term plans or complex future situations, but focuses more on how to achieve local or short-term goals in the current or upcoming steps.

R3M \cite{R3M} proposed the  visual encoder, which uses a large amount of Ego4D \cite{grauman2022around} human video dataset training to learn sparse and compact representations. Ultimately, it can improve the success rate of robot operations by 10\% compared with visual encoders such as CLIP \cite{radford2021learning} and MoCo \cite{he2020momentum}. After improvement, Vi-PRoM \cite{jing2023exploring} focuses on exploring the pre-training strategy of visual encoders from three dimensions: dataset, model architecture, and training method. Its experimental results show that its performance is stronger than R3M \cite{R3M}.

The problem with R3M \cite{R3M} is that the amount of data used is too small. Paper \cite{radosavovic2023real} uses a larger dataset, including Ego4D \cite{grauman2022around}, ImageNet \cite{deng2009imagenet}, Epic Kitchens \cite{damen2022rescaling}, Something Something \cite{goyal2017something} and 100 Days of Hands \cite{shan2020understanding} datasets, totaling 4.5 million images. Paper \cite{radosavovic2023real} designed a visual encoder for robot manipulation. Thanks to the use of a large amount of Internet data and the introduction of the masked autoencoder (MAE) method, the trained visual encoder was finally frozen for use in physical action strategies, which showed an improvement of up to 75\% over CLIP \cite{radford2021learning} and 81\% over supervised pre-training ImageNet \cite{deng2009imagenet}. 

Some researchers \cite{karamcheti2023language, li2023mastering, du2024learning, lynch2023interactive} focus on using a pre-trained visual encoder and a large amount of robot data to train the Transformer model to directly output robot actions, thereby achieving better generalization capabilities of the robot. RT-1 \cite{brohan2022rt} was trained with a large amount of open space data to achieve the robot's multi-task learning ability and generalization ability in unknown scenarios. It also explored the generalization ability of the robot's skills from three aspects: data size, model parameter size, and data diversity. based on RT-1, Google DeepMind introduced a pre-trained visual language model MOO \cite{stone2023open} to process the observed images and then input them into the visual encoder to achieve zero-shot learning of new environments and new objects.Similarly, the Google DeepMind team proposed a method based on RT-1, Q-transformer \cite{chebotar2023q}, which uses Transformer to train and learn the Q value of each action in reinforcement learning instead of directly outputting the action. The results show that it is superior to previous discrete reinforcement learning and imitation learning methods in real-world operation tasks. In order to further improve the task generalization ability of the robot operation model, the Google team introduced a large language model and a new pre-trained visual encoder to form a new robot model RT-2 \cite{brohan2023rt}, and created a new name for this type of model vision-language-action, uses a large amount of Internet data and robot data for training, which significantly improves the generalization ability of objects. The UC Berkeley team proposed a fine-tuning strategy for policy transfer named Octo \cite{team2024octo} between different observations and different robot actions. Testing of robots on nine different platforms showed that this method is an effective fine-tuning deployment strategy and has important guiding significance for the policy design of future general-purpose robots.In order to enable the robot action strategy to be trained on some low-performance platforms, a robot action framework RoboFlamingo \cite{li2023vision} that was fine-tuned based on the open source visual language model OpenFlamingo \cite{awadalla2023openflamingo}. Experimental results show that this method is cost-effective and easy to use. 

Vima \cite{jiang2022vima} is a Transformer-based robot action strategy that introduced visual cropped images of the target into the prompt. Experiments showed that in zero-shot generalization tasks, given the same training data, the task success rate of the proposed scheme was 2.9 times higher than that of other methods. In order to reduce the interference of natural language on the policy action network, RT-H \cite{belkhale2024rt} first predicts the action language and then predicts the robot action based on the action language and visual information. Experiments show that RT-H is more powerful and flexible using this language action hierarchy, can respond to language intervention, and is superior to methods that learn from remote operation intervention.Training a vision-language-action model from scratch is often costly. based on the open source vision-language model prismatic-vlm, the model Openvla \cite{kim2024openvla} contains pre-trained DINOv2 \cite{oquab2023dinov2} and SigLIP \cite{zhai2023sigmoid} visual encoders and the Llama 2 \cite{touvron2023llama} language model, and finally outputs actions. Experimental results show that the task success rate is 16.5\% higher than that of closed models such as RT-2-X, and the number of parameters is reduced by 7 times.

Also based on Transformer, Hiveformer \cite{guhur2023instruction} that integrates natural language instructions, multi-view scene observations, and historical records of actions and observations to output robot actions. Experiments show that this method solves language-conditional instructions, significantly outperforming the current best methods and having excellent generalization performance. 

To simplify the steps, a simple GPT-style end-to-end model named GR-1 \cite{wu2023unleashing} takes language instructions, a series of observed images, and a series of robot states as input. It then predicts robot actions as well as future images in an end-to-end manner. Voxposer \cite{huang2023voxposer} uses the code writing function of a large language model to interact with the visual language model to form a 3D value graph. The combined value graph is used in a model-based planning framework to synthesize robot trajectories with zero samples. Experiments show that it can perform various tasks in natural language.

Some researchers \cite{lynch2020language,du2024learning} regard language instructions as a kind of goal, convert natural language instructions, goal images, and task IDs into implicit expressions of the goals, combine visual perception and proprioception, and use imitation learning to implement language-based robot action strategies. Paper \cite{ha2023scaling} proposed a robot learning framework to implement a multi-task language-conditioned visual-motor strategy. This strategy integrates proprioception and extends the diffusion strategy single-task behavior cloning method to multi-task with language conditions. The results show that the average success rate is improved by 33.2\%.

\subsubsection{Language-based Long-horizon Action Policy}
When making decisions, the agent considers goals and strategies over a longer time span, rather than focusing only on short-term immediate goals. This long-horizon policy considers the longer term future when making decisions, usually involving more complex planning and reasoning, with the goal of maximizing long-term rewards or achieving long-term goals.

Real-world tasks are often complex and time-consuming, so it is often necessary to decompose a complex task into several subtasks to execute. Currently, many researchers \cite{li2022pre, sharma2021skill} focus on introducing new large model methods to achieve the decomposition of complex tasks, namely high-level action policy. SayCan \cite{ahn2022can} connected a large language model with the real world by pre-training skills. The combination of low-level skills and the large language model helped the robot perform complex tasks. The results showed that it was able to complete long-horizon, natural language instructions and complex tasks. Zero-Shot Planners \cite{huang2022language} used the world knowledge learned by a large language model to achieve task planning and decomposition by describing the prompt of the task without additional training. Text2Motion \cite{lin2023text2motion} proposed a language-based planning framework that enables robots to solve long-horizon sequential manipulation tasks, construct task and motion plans given natural language instructions, and verify whether the plan is completed. Text2Motion uses Q functions encoded in a skill library to guide task planning using a large language model. Experiments have shown that it surpasses the most advanced language-based planning methods. The paper \cite{hu2023look} is based on GPT-4V and introduces visual information to help language models better plan tasks. EmbodiedGPT \cite{mu2024embodiedgpt} is a model that embodies intelligent multimodal understanding and reasoning, combining visual and language information to extract task-related features to achieve efficient and accurate task planning. Palm-e \cite{driess2023palm} performed end-to-end training on different tasks based on a pre-trained large model, and the input prompts contained language and image information, ultimately achieving the generalization of visual language models (such as picture-based language question answering, task planning, etc.). TPVQA \cite{zhang2023grounding} used visual language models to detect whether the execution of a task was successful, and at the same time decomposed the task to generate a planning sequence of subtasks, ultimately showing that the solution can relatively well complete complex and long sequence tasks. TaPA \cite{wu2023embodied} proposed a task planning in embodied tasks, which aligns LLMs with a visual perception model and generates a sequence of executable plans based on the objects perceived in the scene. Experimental results show that it achieves a higher task success rate than LLaVA and GPT-3.5. ViLaIn \cite{shirai2024vision} proposed a visual language interpreter, which uses LLMs and a visual language model to generate problem description. ViLaIn receives error message feedback from a symbolic planner to optimize the generated problem description. Experimental results show that ViLaIn can generate grammatically correct questions with an accuracy of over 99\% and an accuracy of over 58\% for effective plans. PG-InstructBLIP \cite{gao2024physically} integrated a physical-world-based LVLM and a large language model to construct an interactive framework for a robot planner. This method showed better planning performance on tasks that require reasoning about physical object concepts. Experimental results show that it can improve the success rate of tasks. 

In addition to other models, GPT can also be used to complete robotics tasks. Model ision \cite{wake2024gpt} used a method to enhance the general visual language model GPT-4V, input demonstration operation videos and natural language instructions, and encoded this information into symbolic task plans based on the GPT-4 task planner. The results showed that it can efficiently learn from demonstration videos.

ScreenAgent \cite{niu2024screenagent} simply using task planning cannot solve complex tasks well, and that it is necessary to further tightly couple task planning and motion planning. Autotamp \cite{chen2024autotamp} converted natural language task descriptions into intermediate task representations, and then used traditional task-and-motion planning algorithms combined with intermediate task representations to jointly solve task and motion planning. Paper \cite{zhou2023generalizable} used the knowledge reasoning ability of LLMs to generate task execution condition descriptions for generalization to new objects and unseen tasks. The task execution condition descriptions guide the generation and adjustment of Dynamic Movement Primitives trajectories to perform long-horizon tasks. Mutex \cite{shah2023mutex} proposed a unified method for policy learning from multi-modal task specifications. The Transformer-based architecture integrates cross-modality and Transformer-based action policy learning, effectively learns cross-modal task specifications, and can execute sequence text task instructions and multi-task text instructions at the same time. The final results show that the performance exceeds that of a single modality. Octopus \cite{yang2025octopus} introduced an embodied visual language programmer, which uses the generated executable code as a medium to connect high-level task planning and real-world operations, and demonstrated the effectiveness of this method through a series of experiments. Some methods like 3d-vla \cite{zhen20243d} and Grid \cite{ni2023grid} believe that 3D scene information can help models better understand the real world. LEO \cite{huang2023embodied} developed an embodied multimodal generalist agent that excels at perceiving, reasoning, planning, and acting in the 3D world. Through 3D visual language alignment and 3D visual language action instruction adjustment, they experimentally demonstrated the method's outstanding capabilities in a wide range of tasks, including question answering, embodied reasoning, task planning, and action strategies. Sayplan \cite{rana2023sayplan} proposed a scalable method to construct robot task planning based on LLMs and 3D scene graph representation. In large-scale environment evaluation, it shows that our method can lay the foundation for robots to execute large-scale, long-term task planning from natural language instructions.

\begin{table}[H]
    \centering
    \caption{Embodied Interaction large models}
    \label{tab:Interaction}
    \setlength{\tabcolsep}{3pt}
    \resizebox{1\columnwidth}{!}{
    \begin{tabular}{c >{\centering\arraybackslash}m{3cm} c >{\centering\arraybackslash}m{4cm} >{\centering\arraybackslash}m{3cm} >{\centering\arraybackslash}p{3cm} c}
        \toprule
        Model & Architecture & Size & Platform & Dataset & Hardware & Year \\
        \toprule
        R3M \cite{R3M} & ResNet \cite{he2016deep} & - & MetaWorld, Franka Kitchen, Adroit, Franka Emika Panda & Ego4D \cite{grauman2022around} & - & 2022 \\
        \hline
        Paper \cite{radosavovic2023real} & ViT \cite{dosovitskiy2020image} & - & - & Ego4D \cite{grauman2022around}, ImageNet \cite{deng2009imagenet}, Epic Kitchens \cite{damen2022rescaling}, Something Something \cite{goyal2017something}, 100 Days of Hands \cite{shan2020understanding} & - & 2022 \\
        \hline
        MOO \cite{stone2023open} & OWL-ViT \cite{minderer2022simple}, RT-1 \cite{brohan2022rt} & - & 7-DOF robotic arm, two-finger gripper mobile manipulation robot & RT-1 \cite{brohan2022rt}, pick & - & 2023 \\
        \hline
        Q-transformer \cite{chebotar2023q} & Universal Sentence Encoder \cite{cer2018universal}, FiLM
        EfficientNet \cite{perez2018film, tan2019efficientnet} & - & 7-DOF robotic arm, two-finger gripper mobile manipulation robot & RT-1 \cite{brohan2022rt} & - & 2023 \\
        \hline
        Octo \cite{team2024octo} & T5-base \cite{raffel2020exploring}, ViT \cite{dosovitskiy2020image} & Octo-Small: 27M Octo-Base: 93M & WidowX 250 6-DOF, UR5,  RT-1 Robot, ALOHA, Trossen ViperX & Open X-Embodiment \cite{o2024open} & - & 2024 \\
        \hline
        Vima \cite{jiang2022vima} & T5, Mask R-CNN \cite{he2017mask} & 2M-200M &  & VIMA-BENCH built on the Ravens simulator \cite{zeng2021transporter, shridhar2022cliport} & - & 2022 \\
        \hline
        Hiveformer \cite{guhur2023instruction} & CLIP, UNet \cite{ronneberger2015u} & - & 6-DOF UR5 robotic arm with 2-finger Robotiq RG2 gripper and two cameras & real, RLBench \cite{james2020rlbench} & NVIDIA Tesla V100 SXM2 GPU & 2023 \\
        \hline
        GR-1 \cite{wu2023unleashing} & CLIP, ViT & 195M & Kinova Gen2 & Ego4D \cite{grauman2022around}, CALVIN \cite{mees2022calvin}, real & - & 2023 \\
        \hline
        Paper \cite{ha2023scaling} & CLIP & - & 6-DOF Robotic Arm & Based on MuJoCo simulator \cite{todorov2012mujoco} and Google Scanned dataset \cite{downs2022google} & NVIDIA RTX 3080 GPU & 2023 \\
        \hline
        SayCan \cite{ahn2022can} & LLM, Skills Library, Value Functions, Decision Module & - & mobile manipulator & 68,000 data points collected through VR devices, RetinaGAN \cite{ho2021retinagan} & 16 TPUv3 and 3000 CPU & 2022 \\
        \hline
        Zero-Shot Planners \cite{huang2022language} & GPT-3 \cite{GPT3}, Codex 12B \cite{chen2021evaluating} & 175B/12B & - & VirtualHome \cite{puig2018virtualhome}, ActivityPrograms \cite{puig2018virtualhome} & - & 2022 \\
        TaPA \cite{wu2023embodied} & LLaMA-7B, open-vocabulary object detector & - & - & AI2-THOR \cite{kolve2017ai2} and real & 8 GTX 3090 GPU & 2023 \\
        ViLaIn \cite{shirai2024vision} & Grounding-DINO \cite{garrett2020pddlstream}, GPT-4 \cite{GPT4}, BLIP-2 \cite{li2023blip} & - & - & ProDG \cite{shirai2024vision} & - & 2024 \\
        \hline
        PG-InstructBLIP \cite{gao2024physically} & InstructBLIP \cite{dai2023instructblip}, fine-tuning in PHYSOBJECTS \cite{gao2024physically} & 11B/11B & Franka Emika Panda & PHYSOBJECTS \cite{gao2024physically}, real & - & 2024 \\
        \bottomrule
    \end{tabular}
    }
\end{table}
\begin{table}[H]
    \centering
    \label{tab:Interaction_1}
    \setlength{\tabcolsep}{3pt}
    \resizebox{1\columnwidth}{!}{
    \begin{tabular}{c >{\centering\arraybackslash}m{3cm} c >{\centering\arraybackslash}m{4cm} >{\centering\arraybackslash}m{3cm} >{\centering\arraybackslash}p{3cm} c}
        \hline
        Model & Architecture & Size & Platform & Dataset & Hardware & Year \\
        \hline
        Text2Motion \cite{lin2023text2motion} & LLM, Library of Learned Skills, Geometric Feasibility Planner, Hybrid Planning Algorithm & - & Franka Panda, Kinect V2 & 1 million skill training datasets and OOD calibration datasets & Nvidia Quadro P5000 GPU, 2 CPU & 2023 \\
        \hline
        EmbodiedGPT \cite{mu2024embodiedgpt} & ViT-G/14 \cite{fang2023eva}, LLaMA-7B \cite{touvron2023llama} & 10B & Franka Emika Panda & EgoCOT, EgoVQA, COCO Caption \cite{lin2014microsoft}, CC3M \cite{sharma2018conceptual}, LAION-400M & - & 2024 \\
        Palm-e \cite{driess2023palm} & LLM, ViT & 12B, 84B, 562B & Franka Emika Panda & TAMP, Language-Table \cite{lynch2023interactive}, Mobile Manipulation, WebLI \cite{chen2022pali}, VQA v2 \cite{goyal2017making}, COCO \cite{chen2015microsoft}, OK-VQA \cite{marino2019ok} and others & - & 2023 \\
        \hline
        ision \cite{wake2024gpt} & GPT-4V & - & Nextage, Fetch Mobile Manipulator, SEED-noid, Shadow Dexterous Hand Lite & Cooking Video dataset & - & 2024 \\
        \hline
        ScreenAgent \cite{niu2024screenagent} & CogAgent \cite{hong2024cogagent} & - & - & ScreenAgent \cite{niu2024screenagent}, COCO \cite{chen2015microsoft}, Widget Captions \cite{li2020widget}, Mind2Web \cite{deng2024mind2web} and others & - & 2024 \\
        \hline
        Autotamp \cite{chen2024autotamp} & LLM, TAMP & - & Differential-Drive Robots and Virtual environment robot & HouseWorld1 \cite{finucane2010ltlmop}, HouseWorld2, Chip’s Challenge, Overcooked, Rover \cite{sun2022multi}, Wall \cite{sun2022multi} & - & 2024 \\
        \hline
        AURL \cite{thankaraj2023sounds} & ResNet \cite{he2016deep} &  -& UR10 & self-collected dataset & - & 2023 \\
        \hline
        Maniwav \cite{thankaraj2023sounds} & CLIP, AST \cite{gong2021ast} & - & UR5 robot arm & self-collected dataset & - & 2024 \\
        \hline
        LEO \cite{huang2023embodied} & Vicuna-7B \cite{chiang2023vicuna}, LoRA \cite{hu2021lora} & 7B & - & LEO-align \cite{huang2023embodied}, LEO-instruct \cite{huang2023embodied} & 4 $\times$ NVIDIA A100 GPU & 2023 \\
        \hline
        Paper \cite{zhou2023generalizable} & LLM, Dynamic Movement Primitives & - & Pybullet, MOVO & - & - & 2023 \\
        \hline
        Mutex \cite{shah2023mutex} & ViT-L/14 \cite{radford2021learning}, Whisper-Small \cite{radford2023robust} & - & LIBERO-100 \cite{liu2024libero} and 50 real-world tasks & LIBERO-100 \cite{liu2024libero}, real & 2 $\times$ NVIDIA RTX A5000 (24 GB) GPU & 2023 \\

        \hline
    \end{tabular}
    }
\end{table}

\begin{table}[t!]
    \centering
    \label{tab:Interaction_2}
    \setlength{\tabcolsep}{3pt}
    \resizebox{1\columnwidth}{!}{
    \begin{tabular}{c >{\centering\arraybackslash}m{3cm} c >{\centering\arraybackslash}m{4cm} >{\centering\arraybackslash}m{3cm} >{\centering\arraybackslash}p{3cm} c}
        \hline
        Model & Architecture & Size & Platform & Dataset & Hardware & Year \\
        \hline
        TPVQA \cite{zhang2023grounding} & Planning Domain Definition Language (PDDL) or Answer Set Programming (ASP), Visual Question Answering(VQA), ViLBERT \cite{lu2019vilbert} & - & UR5e robotic arm with Hand-E gripper mounted on a Segway base and equipped with an RGB-D camera, DALL-E \cite{ramesh2021zero} based simulator & VQA v2.0 \cite{goyal2017making} & - & 2023 \\
        \addlinespace[0.5em]
        3d-vla \cite{zhen20243d} & 3D-LLM \cite{hong20233d}, BLIP2-FlanT5XL \cite{li2023blip} & - & - & 3D Embodied Instruction Tuning dataset \cite{zhen20243d} & 6 $\times$ 32 V100 GPU, 6 $\times$ 64 V100 GPU & 2024 \\
        \hline
        Grid \cite{ni2023grid} & LLM (INSTRUCTOR) , Graph Attention Networks, cross-attention, Task Decoder & - & Unity, real & Self-built dataset \cite{ni2023grid} & 2 $\times$ NVIDIA RTX 4090 GPU & 2023 \\
        \hline
        Sayplan \cite{rana2023sayplan} & GPT-4, 3DSG & - & Franka Panda 7-DoF, Omron LD-60, LiDAR & Office Environment, Home Environment & - & 2023 \\
        \hline
    \end{tabular}
    }
\end{table}

\subsection{Simulation}

\begin{table}[t!]
    \centering
    \caption{Simulators for Embodied Multimodal Large Models}
    \label{tab:Simulators}
    \setlength{\tabcolsep}{3pt}
    \resizebox{1\columnwidth}{!}{
    \begin{tabular}{c >{\centering\arraybackslash}m{3cm} >{\centering\arraybackslash}m{4cm} >{\centering\arraybackslash}m{3cm} >{\centering\arraybackslash}m{3cm} c}
        \hline
        Name & Task & Scenes & Sensors & Platform & Year \\
        \hline
        SDF-Sim \cite{rubanova2024learning} & Action, Navigation & - & RGB & - & 2024\\
        TRUMANS \cite{jiang2024scaling} & Action, Navigation & 100 interior scenes such as dining room, living room, bedroom and kitchen, etc. &  VICON, RGB-D, IMU & A800 GPU & 2024\\
        \addlinespace[0.5em]
        WonderWorld \cite{yu2024wonderworld} & Action, Navigation & - & - & A6000 GPU, AR, VR & 2024\\
        \addlinespace[0.5em]
        GenZI \cite{li2024genzi} & Action, Navigation & - & - & A100 GPU & 2024\\
        \hline
        iGibson2.0 \cite{iGibson2.0} & Action, Navigation & 15 complete interactive scenes with 108 rooms such as kitchen, bathroom, living room, etc. & RGB, D, LiDAR & Intel 5930k CPU, Nvidia GTX 1080 Ti GPU,  HTC Vive (Pro Eye), Oculus Rift S, Oculus Quest, Fetch robot, Humanoid robot with two arms & 2021\\
        \hline
        Habitat-Sim \cite{savva2019habitat} & Action, Navigation & - & RGB, D, GPS, Compass, Contact & Intel Xeon E5-2690 v4 CPU, Nvidia Titan Xp GPU, VR, Robot & 2019\\
        \hline
        Genesis \cite{Genesis} & all & all & - & - & 2024\\
        \hline
        Matterport3D \cite{chang2017matterport3d} & Navigation & 90 architectural-scale scenes such as homes, offices, and churches & RGB-D, Panoramic & Intel Xeon E5-2690 v4 CPU, Nvidia Titan Xp GPU, Robot & 2017\\
        
        SoundSpaces \cite{chen2020soundspaces} & Navigation & 103 scenes, 102 copyright-free sounds & RGB-D, microphone & - & 2020\\
        
        SoundSpaces v2 \cite{chen2022soundspaces} & Navigation & - &  RGB-D, microphone & - & 2022\\
        \hline
        
    \end{tabular}
    }
\end{table}

Embodied simulation is essential for embodied intelligence as it allows for the design of precisely controlled conditions and optimizes the training process. This enables agents to be tested in various environmental settings, enhancing their understanding and interaction capabilities. Additionally, it fosters cross-modal learning within the embodied agents themselves and facilitates the training and evaluation of generated data. To enable meaningful interaction with the environment, a realistic simulation environment must be constructed, taking into account the physical characteristics of the surroundings, the properties of objects, and their interactions. Simulation platforms generally fall into two categories: general simulators based on foundational simulations and simulators based on real-world scenarios. \cite{liu2024aligning}. Some of the latest and famous simulation methods and platforms are shown in Table \ref{tab:Simulators}.

\subsubsection{General Simulators Based on Foundational simulations}
NVIDIA Omniverse™ Isaac Sim is a powerful robotics simulation toolkit designed for the NVIDIA Omniverse™ platform. It equips researchers and practitioners with essential tools and workflows to build virtual robotic environments and conduct experiments. Isaac Sim enables the creation of highly accurate, physically realistic simulations and synthetic datasets, offering capabilities such as advanced physics simulation and multi-sensor RTX rendering. With support for ROS2, Isaac Sim facilitates the design, debugging, training, and deployment of robots, helping to accelerate the development of autonomous systems.

\subsubsection{Simulators Based on Real-world Scenarios}
In response to the challenges faced by traditional physics simulators in handling large-scale scenes, the learning-based rigid body simulator SDF-Sim \cite{rubanova2024learning} leverages learned signed distance functions (SDFs) to represent object shapes. This approach aims to accelerate distance calculations and enable efficient simulation of complex, large-scale environments.

Paper \cite{jiang2024scaling} introduces the TRUMANS \cite{jiang2024scaling} dataset and proposes an autoregressive motion diffusion model for generating human-scene interaction (HSI) sequences. The model incorporates a local scene sensor and a frame-by-frame action embedding module. The local scene sensor captures the contextual information of the scene, while the action embedding module processes action labels on a frame-by-frame basis. Together, these components enhance the model's ability to understand and control both the scene context and the actions, improving the generation of realistic HSI sequences.

3D scene generation has garnered significant attention, but most existing methods rely on offline generation, which often leads to issues such as slow generation speeds and scene geometry distortion. These limitations hinder their ability to support real-time interaction and diverse scene creation, which are essential for applications like game development and virtual reality (VR). The WonderWorld \cite{yu2024wonderworld} framework addresses these challenges by enabling the generation of diverse and coherent 3D scenes from a single image, achieving low-latency user interaction. It employs a fast Layered Gaussian Facet (FLAGS) representation, where the 3D scene is modeled as a radiation field consisting of three layers: foreground, background, and sky. Each layer is made up of a set of facets. Using a single-view layer generation approach, the framework generates images and masks for each layer from a single scene image. This method results in higher-quality scenes, improved semantic alignment, better consistency across novel views, and faster generation speeds.

GenZI \cite{li2024genzi} is a pioneering zero-shot method designed to generate 3D human-scene interactions (HSI) from text descriptions, which addresses the challenge of generating 3D interactions without relying on annotated 3D data. This method automatically generates human figures in multiple rendered views of a 3D scene and uses a LVLM to generate potential 2D interaction hypotheses. These 2D hypotheses are then transformed into 3D representations by optimizing the pose and shape parameters of a 3D human model (SMPL-X). By bridging the gap between text-based descriptions and 3D human-scene interactions, GenZI provides a powerful tool for creating realistic 3D HSIs without the need for extensive 3D datasets.

The approach used in GenZI can be related to simulation environments like iGibson 2.0 \cite{iGibson2.0} and Habitat-Sim \cite{savva2019habitat}, which focus on generating interactive 3D environments for embodied AI research. iGibson 2.0 extends multiple states for objects (e.g., temperature, humidity, cleanliness) and defines logical predicates to simulate a variety of household tasks. It is also compatible with commercial VR systems, enabling users to interact within virtual scenes. This system shares a common goal with GenZI in that both methods aim to create interactive and dynamic 3D environments, with iGibson focusing on environmental realism and task simulation, while GenZI targets realistic human interactions with scenes based on textual descriptions.

Similarly, Habitat-Sim \cite{savva2019habitat} provides a high-performance 3D simulation environment tailored for embodied AI research, with support for various 3D scenes, configurable sensors, and robot models. Habitat-Sim's efficient simulation capabilities (with rendering speeds reaching over 10,000 FPS) make it a powerful platform for large-scale AI agent training. While iGibson 2.0 focuses on simulating household environments with extended object states, Habitat-Sim supports more general-purpose 3D scene generation and robot interaction, which is critical for developing and testing AI agents in diverse environments. Both Habitat-Sim and iGibson enable users to customize physical parameters and robot models, offering flexibility in creating dynamic scenarios for AI research.

Together, these methods GenZI, iGibson 2.0, and Habitat-Sim share the common objective of improving the realism and interactivity of 3D simulations, each contributing a unique approach. GenZI focuses on transforming text-based descriptions into 3D human-scene interactions, while iGibson 2.0 and Habitat-Sim provide the virtual environments necessary for training embodied AI agents. The synergy between these methods illustrates how different techniques can complement each other to push the boundaries of realistic and interactive simulations for AI research.

It is important to emphasize that the latest technology Genesis \cite{Genesis} is powered by a newly developed general physics engine. This engine integrates a variety of physics solvers and their interactions into a unified framework. Built on this core engine, a generative agent framework has been introduced to enable fully automated data generation for robotics and other domains. This framework supports a wide range of modalities, including physically accurate and spatially consistent video, camera motion and parameters, actions of human and animal characters, robotic manipulation and motion strategies designed for real-world deployment, fully interactive 3D environments, open-world articulated object generation, as well as voice audio, facial expressions, and actions.
\section{Datasets}
\label{sec:DATA}
In this section, we first describe the dataset collection method, followed by an introduction to the datasets used for perception and interaction models, as well as the datasets used for navigation models.

\subsection{Embodied Datasets Collection Methods}

There are two primary methods for collecting datasets related to embodied intelligence: one involves using an intelligent agent with a physical body to gather data in the real world, while the other relies on collecting datasets through a simulator.

The dataset, similar to those in  \cite{o2024open, wang2024all}, was collected in a real-world environment using various sensors, including RGB cameras, depth cameras, IMUs, LiDAR, pressure sensors, sound sensors, and others. However, during the data collection process, issues such as occlusions in the field of view or incomplete recording of operational details may arise. To address these challenges, DexCap  \cite{wang2024dexcap} utilizes Simultaneous Localization and Mapping (SLAM) to track hand movements.

Another type of dataset is collected using simulators, such as Unity and Gazebo. This approach enables the rapid generation of large volumes of multimodal data (e.g., images, depth maps, sensor data, etc.) while offering control over environmental and task variables, facilitating model training. Some of the latest and most widely used simulators are listed in Table \ref{tab:Simulators}.

\subsection{Embodied Perception and Interaction Datasets}

Several recent datasets have played a pivotal role in advancing the development of embodied intelligence for robots.

Notably, the Open X-Embodiment Dataset \cite{o2024open}, released by the Google team in collaboration with over 20 organizations and research institutes, provides a large-scale multi-modal resource. It includes data from 22 types of robots, capturing RGB images, end-point motion trajectories, and language commands across 1 million scenes, 500+ skills, and 150,000 tasks. It contains 60 datasets, some of which are shown in Table \ref{tab:dataset}.

The field of embodied intelligence relies heavily on diverse datasets that capture various robot operations, environments, and sensory modalities. These datasets can generally be categorized based on their data collection methods, such as real-world data, simulated data, or a combination of both, with several datasets incorporating multimodal information.

One example is the RH20T dataset  \cite{fang2024rh20t}, introduced by Hao-Shu Fang et al., which comprises over 110,000 robot operation sequences. This dataset offers a range of data modalities, including vision, force, audio, motion trajectories, demonstration videos, and natural language instructions, making it a valuable resource for training embodied intelligence models. Similarly, the ManiWAV dataset  \cite{liu2024maniwav} uses an 'ear-in-hand' data collection device to capture human demonstrations in real-world settings. It synchronizes audio and visual feedback, offering a rich source of data for learning robot manipulation policies directly from human demonstrations.

A large-scale example of multimodal data is the All Robots in One (ARIO) dataset  \cite{wang2024all}, developed by Peng Cheng Laboratory. With over 3 million samples, the ARIO dataset includes images, language commands, tactile feedback, and speech from a variety of robot platforms. It spans data collected in real-world scenarios using platforms like the Cobot Magic and Cloud Ginger XR-1 as well as data generated in simulation platforms such as Habitat \cite{savva2019habitat}, MuJoCo \cite{todorov2012mujoco}, and SeaWave \cite{ren2024surferprogressivereasoningworld}. It also includes converted data from other datasets like Open X-Embodiment \cite{o2024open}, RH20T \cite{fang2024rh20t}, and ManiWAV \cite{liu2024maniwav}, further enriching its multimodal scope.

Simulated environments also play a crucial role in data collection, as seen in the ManiSkill  \cite{mu2021maniskill} and ManiSkill2  \cite{gu2023maniskill2} datasets from UC San Diego. These datasets contain 36,000 successful manipulation trajectories along with 1.5 million point clouds and RGB-D frames, all captured in simulation. Similarly, the 3D-VLA  \cite{zhen20243d} dataset includes robot data such as 2D datasets from Open-X Embodiment \cite{o2024open}, depth-inclusive datasets like Dobb-E \cite{shafiullah2023dobbe} and RH20T \cite{fang2024rh20t}, and simulation datasets like RLBench \cite{james2020rlbench} and CALVIN \cite{mees2022calvin}. It also incorporates human-object interaction data, including the HOI4D~\cite{Liu_2022_CVPR} dataset.

Other notable datasets that blend real-world and simulated data include the DROID Dataset  \cite{khazatsky2024droid}, which contains 76,000 demonstration trajectories, 564 scenes, and 86 tasks with multi-modal data, and the BridgeData V2  \cite{walke2023bridgedata}, which offers 60,096 robot trajectory samples from 24 environments. Both datasets integrate sensory modalities such as images, depth, and natural language commands.

In addition to manipulation tasks, some datasets focus on more specialized areas of robot control. The TACO-RL  \cite{rosete2022tacorl} dataset, for example, is designed for training hierarchical policies to solve long-term robot control tasks by teleoperating robots in simulated and real environments. Meanwhile, FurnitureBench  \cite{heo2023furniturebench} is focused on testing complex, long-term operational tasks related to furniture assembly, emphasizing skills such as precise grasping, path planning, and insertion.

Finally, datasets like the Dexterous Hand dataset  \cite{fan2023arctic}, introduced by ETH Zurich, contain 2.1 million video frames, 3D hand and object meshes, dynamic contact information, and hand posture and object state trajectories, providing valuable insights into dexterous hand manipulation. The CLVR Jaco Play dataset  \cite{dass2023jacoplay} from USC, which includes 1,085 teleoperated robot episodes with various data modalities, is another key resource for training robots in manipulation tasks.

In summary, these diverse datasets, ranging from real-world to simulated environments, offer a wealth of multimodal data that enable advancements in embodied intelligence, robotic manipulation, and human-robot interaction.

For datasets specifically targeting touch-related modalities, notable contributions include the TVL dataset \cite{fu2024touch} and the Touch 1k dataset \cite{cheng2024touch100k}, which focus primarily on multi-modal alignment or representation learning. A more comprehensive list of such datasets can be found in Touch100k \cite{cheng2024touch100k}. However, few of these datasets have been applied directly to robotic tasks, such as interaction or navigation. While smaller-scale datasets like those discussed in  \cite{calandra2018more} and  \cite{han2024learning} are available for specific tasks, they are not designed for learning general models and have limited applicability.

\subsection{Embodied Navigation Datasets}
Embodied navigation datasets aim to enhance robots' ability to accurately navigate in physical or simulated environments based on visually-linguistic combined instructions. This is achieved by providing long and complex paths and instructions, real-world data, diverse indoor and outdoor scenes, support for training large high-capacity models, and detailed intermediate products such as 3D scene reconstructions, relative depth estimates, object labels, and localization information. These datasets effectively expand the application scenarios of vision-language navigation and provide strong data support for solving practical downstream application problems. The dataset is shown in Table \ref{tab:dataset}.

HM3D \cite{ramakrishnan2021habitat} is the largest-scale building-scale 3D scene reconstruction dataset, containing diverse real spaces from around the world. HM3D provides 1,000 nearly complete high-fidelity reconstructions of entire buildings. Each reconstruction was captured using the Matterport Pro2 tripod-based depth sensor to capture the habitable and navigable spaces of each interior space. Additionally, the HM3D dataset seamlessly integrates with FAIR's Habitat simulator, supporting the training and evaluation of agents (such as home robots and AI assistants).

The Gibson environment \cite{xia2018gibson} designed for training and testing real-world perceptive agents. Gibson is based on virtualized real spaces, rather than artificially designed spaces, and currently includes over 1400 floor spaces from 572 complete buildings. The main features of Gibson include: I. Originating from the real world and reflecting its semantic complexity, II. The internal synthesis mechanism ``Goggles'' allows trained models to be deployed in the real world without the need for domain adaptation, III. The embodiment of the agent and its subjection to physical and spatial constraints.

The Matterport3D dataset \cite{chang2017matterport3d} is a large-scale RGB-D dataset, containing 10,800 panoramic views and 194,400 RGB-D images from 90 building-scale scenes. The dataset provides annotations for surface reconstruction, camera poses, as well as 2D and 3D semantic segmentation. Based on the Matterport3D environment, Peter Anderson et al. collected the R2R dataset \cite{anderson2018vision}, which includes 21,567 open-vocabulary, crowdsourced navigation instructions with an average length of 29 words. Each instruction describes a trajectory that typically crosses multiple rooms. The associated task requires the agent to navigate to a target location in a previously unseen building by following natural language instructions. Like R2R, REVERIE is built on the Matterport3D simulator, providing detailed indoor navigation environments. Unlike R2R, REVERIE dataset \cite{Qi_2020_CVPR} introduces the Remote Embodied Visual Referring Expression in Real Indoor Environments task for the first time, requiring the agent to navigate and identify remote objects in real indoor environments based on natural language instructions. The target object is not observable from the starting point, which means the robot must have common sense and reasoning abilities to reach the location where the target might appear. The goal is to assess the agent's ability to navigate and identify target objects based on advanced natural language instructions, emphasizing the robot's need for natural language understanding and visual navigation.

The ScanQA dataset \cite{Azuma_2022_CVPR} is designed for a 3D question-answering (3D-QA) task in three-dimensional spatial scene understanding. It requires models to receive visual information from the entire 3D scene derived from rich RGB-D indoor scans and to answer given textual questions about that 3D scene. The dataset contains over 41k question-answer pairs from 800 indoor scenes in the ScanNet dataset. These question-answer pairs are manually editeD,  with answers associated with the 3D objects in each 3D scene. The ScanQA dataset aims to advance research in 3D spatial understanding by answering questions about 3D scenes through the combination of linguistic expressions and 3D geometric features.

The LLaVA dataset \cite{liu2024visual} is a specialized instruction-tuning dataset for multimodal tasks, containing 158K unique language-image instruction-following samples, which are divided into 58K dialogues, 23K detailed descriptions, and 77K complex reasonings. This dataset enhances the model's zero-shot capability on new tasks through visual instruction tuning, a process that enables the model to better understand and execute visual instructions. In navigation applications, the LLaVA dataset can be used to train models to understand and execute instructions based on visual and textual information. This capability is crucial for embodied navigation tasks, which often involve navigating based on visual information. Specifically, the LLaVA dataset supports the model's navigation and execution capabilities in complex environments by providing a diverse range of instruction-tuning data.

Targeting at the audio-assited navigation task, the SoundSpaces simulator~\cite{chen2020soundspaces}, based on Matterport3D \cite{chang2017matterport3d} and Replica \cite{straub2019replica}, takes the space and material property into account to render realistic sound in a 3D space. It includes 85 scenes from Matterport3D,  18 scenes from Replica and 102 copyright-free sounds. To extend the SoundSpaces simulator, the SoundSpaces-v2~\cite{chen2022soundspaces} was designed with ability of generalizing to new environments, freely adjusting the material and microphone configuration.

\begin{table}[H]
    \centering
    \caption{Embodiment Dataset}
    \label{tab:dataset}
    \setlength{\tabcolsep}{3pt}
    \resizebox{1\columnwidth}{!}{
    \begin{tabular}{c >{\centering\arraybackslash}m{3cm} >{\centering\arraybackslash}m{3.5cm} >{\centering\arraybackslash}m{4cm} >{\centering\arraybackslash}m{3cm} >{\centering\arraybackslash}m{3cm} >{\centering\arraybackslash}m{3cm} c}
        \hline
        Dataset & Task & Scenes & Robot & Robot Morphology & Sensors (D: Depth camera. W: Wrist camera) & Language Annotations & Year\\
        \hline
        RT-1 Robot Action \cite{brohan2022rt} & Action & 130k+ episodes,700+ tasks &  Google Robot & Mobile Manipulator & RGB, D & Templated & 2022  \\
        \hline
        QT-Opt \cite{kalashnikov2018qt} & Action & 580k real-world grasp  & Kuka iiwa & Single Arm & RGB & None &  2018 \\
        \hline
        Berkeley Bridge \cite{walke2023bridgedata} & Action & 13 skills, 24 environments, 60,096 trajectories & WidowX & Single Arm & RGB, D, W & Natural & 2023 \\
        \hline
        Freiburg Franka Play \cite{rosete2022tacorl} & Action & - & Franka & Single Arm & RGB,  D, W & Templated & 2022\\
        USC Jaco Play \cite{dass2023jacoplay} & Action & 1085 teleoperated robot episodes & Jaco 2 & Single Arm & RGB,  W & Templated & 2023\\
        Berkeley Cable Routing \cite{luo2023multistage} & Action & 1442 trajectories, 257 tasks & Franka & Single Arm & RGB,  W & None &  2023 \\
        \hline
        Roboturk \cite{mandlekar2019scaling} & Action & three real world tasks,2144 demonstrations & Sawyer & Single Arm & RGB,  D & Templated & 2019  \\
        \hline
        NYU VINN \cite{pari2021surprising} & Action & 435 episodes & Hello Stretch & Mobile Manipulator & RGB,  W & None & 2021 \\
        
        RECON \cite{shah2021rapid} & Navigation  & 11830 episodes & Jackal & Wheeled Robot & RGB, D, W & None & 2021 \\
        HM3D \cite{ramakrishnan2021habitat} & Navigation & 1,000 building-scale scenes & Simulated & Virtual Agent & RGB, D & None & 2021 \\
        Gibson \cite{xia2018gibson} & Navigation & 571 scenes & Simulated & Virtual Agent & RGB, D & None & 2018 \\
        \hline
        R2R \cite{anderson2018vision} & Navigation & 90 buildings & Simulated & Virtual Agent & RGB, D & Natural language instructions & 2018 \\
        \hline
        MP3D \cite{chang2017matterport3d} & Navigation & 90 buildings & Simulated & Virtual Agent & RGB, D & None & 2017 \\
        AI2THOR \cite{kolve2017ai2} & Navigation & 120 scenes  & Simulated & Virtual Agent & RGB, D & None & 2017 \\
        iTHOR \cite{weihs2021visual} & Navigation & 120 rooms & Simulated & Virtual Agent & RGB, D & None & 2022 \\
        \hline
        RxR \cite{ku2020room} & Navigation & 90 houses & Simulated & Virtual Agent & RGB, D & Natural Language Instructions & 2020 \\
        \hline
        R4R \cite{jain2019stay} & Navigation & 90 houses & Simulated & Virtual Agent & RGB, D & Natural Language Instructions & 2019 \\
        \hline
        REVERIE \cite{Qi_2020_CVPR} & Navigation & 90 buildings & Simulated & Virtual Agent & RGB, D & Natural language instructions & 2020 \\
        \hline
        SOON \cite{zhu2021soon} & Navigation & 90 houses & Simulated & Virtual Agent & RGB, D & Natural Language Instructions & 2021 \\
        \hline
        RoboTHOR \cite{deitke2020robothor} & Navigation & 89 scenes & LoCoBot & Wheeled Robot & RGB, D & None & 2020 \\
        \hline
        CVDN \cite{thomason2020vision} & Navigation & 83 houses & Simulated & Virtual Agent & RGB & human dialogs & 2020 \\
        \hline
        ScanQA \cite{Azuma_2022_CVPR} & 3D Question Answering & 800 rooms & Simulated & - & RGB, D & 41k QA pairs & 2022 \\
        LLaVA-23k \cite{liu2024visual} & Visual Instruction Tuning & 80k images & Simulated & - & RGB & 158k instructions & 2023 \\
        \hline
        MP3D-EQA \cite{wijmans2019embodied} & Embodied QA & 83 homes, 144 floors & Simulated & Virtual Agent & RGB, D & 1.1k Template-based QA & 2019 \\  
        \hline
        Austin VIOLA \cite{zhu2022viola} & Action & 1 scene, multi-task household operations & Franka & Single Arm & RGB, W & Templated & 2022 \\
        Berkeley Autolab UR5 \cite{BerkeleyUR5Website} & Action & 4 scenes, 4 tasks & UR5 & Single Arm & RGB, D, W & Templated & \\
        \hline
        TOTO Benchmark \cite{zhou2023train} & Action & 1 scene, 2 tasks (scooping, pouring) & Franka & Single Arm & RGB & None & 2023\\
        \hline
        Language Table \cite{lynch2023interactive} & Action & Multi-scene, natural language instruction tasks, 442k trajectories & xArm & Single Arm & RGB & Natural & 2023\\
        \hline
        Columbia PushT Dataset \cite{chi2023diffusionpolicy} & Action & 1 scene, 2 tasks (T-block pushing), 122 trajectories & UR5 & Single Arm & RGB, W & None & 2023\\
        \hline
        Stanford Kuka Multimodal \cite{lee2019icra} & Action & 1 scene, multi-task (peg insertion), 3k trajectories & Kuka iiwa & Single Arm & RGB & None & 2019\\
        \hline
    \end{tabular}
    }
\end{table}
\begin{table}[t!]
    \centering
    \label{tab:dataset_1}
    \setlength{\tabcolsep}{3pt}
    \resizebox{1\columnwidth}{!}{
    \begin{tabular}{c >{\centering\arraybackslash}m{3cm} >{\centering\arraybackslash}m{3.5cm} >{\centering\arraybackslash}m{4cm} >{\centering\arraybackslash}m{3cm} >{\centering\arraybackslash}m{3cm} >{\centering\arraybackslash}m{3cm} c}
        \hline
        Dataset & Task & Scenes & Robot & Robot Morphology & Sensors (D: Depth camera. W: Wrist camera) & Language Annotations & Year\\
        \hline
        NYU ROT \cite{haldar2023watch} & Action & Multi-scene, diverse manipulation tasks & xArm & Single Arm & RGB & Templated & 2023\\
        \addlinespace[0.5em]
        Stanford HYDRA \cite{belkhale2023hydra} & Action & 3 scenes (kitchen), 3 tasks (coffee, toasting, utensil sorting), 550 trajectories & Franka & Single Arm & RGB, W & Templated & 2023\\
        \addlinespace[0.5em]
        Austin BUDS \cite{zhu2022bottom} & Action & 1 scene, long-horizon kitchen tasks, 50 trajectories & Franka & Single Arm & RGB, W & None & 2022\\
        NYU Franka Play \cite{cui2022play} & Action & 1 scene (toy kitchen), diverse tasks, 456 trajectories & Franka & Single Arm & RGB, D & None & 2022\\
        \hline
        Maniskill \cite{gu2023maniskill2} & Action & Multi-scene (Table Top/Ground), 12 tasks (pick, stack, insert), 30k trajectories & Franka & Single Arm & RGB, D, W & Templated & 2023\\
        \hline
        Furniture Bench \cite{heo2023furniturebench} & Action & Multi-scene, 9 furniture models (assembly tasks), 5,100 trajectories & Franka & Single Arm & RGB, W & Templated & 2023\\
        CMU Franka Exploration \cite{mendonca2023structured} & Action & 1 scene (toy kitchen), exploration tasks & Franka & Single Arm & RGB & Templated & 2023\\
        UCSD Kitchen & Action & Multi-scene (kitchen), complex object manipulation & xArm & Single Arm & RGB & Natural & 2023\\
        \hline
        UCSD Pick Place \cite{Feng2023Finetuning} & Action & 1 scene, pick-and-place with distractors, 1,355 trajectories & xArm & Single Arm & RGB & Templated & 2023\\
        Austin Sailor \cite{nasiriany2022sailor} & Action & 1 scene (toy kitchen), food/utensil manipulation, 250 trajectories & Franka & Single Arm & RGB, W & None & 2022\\
        Austin Sirius \cite{liu2022robot} & Action & 2 scenes, kcup/gear insertion tasks, 600 trajectories & Franka & Single Arm & RGB, W & None & 2023\\
        \hline
        BC-Z \cite{jang2021bc} & Action & Multi-scene (household/office), 100+ tasks, 39k trajectories & Google Robot & Mobile Manipulator & RGB & Templated & 2021\\
        USC Cloth Sim \cite{salhotra2022dmfd} & Action & 1 scene, cloth manipulation, 1k trajectories & Franka & Single Arm & RGB & None & 2022\\
        Tokyo PR2 Fridge Opening \cite{oh2023pr2utokyodatasets} & Action & 1 scene (kitchen), fridge opening tasks & PR2 & Single Arm & RGB & None & 2023\\
        Tokyo PR2 Tabletop Manipulation \cite{oh2023pr2utokyodatasets} & Action & 1 scene, bread/grape manipulation, 192 trajectories & PR2 & Single Arm & RGB & None & 2023\\
        Saytap \cite{saytap2023} & Action & 1 scene (ground), quadrupedal locomotion tasks & Unitree A1 & Quadrupedal Robot & - & Natural & 2023\\
        \addlinespace[0.5em]
        UTokyo xArm PickPlace \cite{matsushima2023weblab} & Action & 1 scene, plate stacking tasks, 95 trajectories & xArm & Single Arm & RGB, W & None & 2023\\
        \addlinespace[0.5em]
        UTokyo xArm Bimanual \cite{matsushima2023weblab} & Action & 1 scene, towel manipulation, 70 trajectories & xArm Bimanual & Bi-Manual & RGB & None & 2023\\
        \hline
        Robonet \cite{dasari2019robonet} & Action & Multi-scene, object interaction, 82k trajectories & Multi-Robot & Single Arm & RGB & None & 2019\\
        \hline
    \end{tabular}
    }
\end{table}
\begin{table}[t!]
    \centering
    \label{tab:dataset_2}
    \setlength{\tabcolsep}{3pt}
    \resizebox{1\columnwidth}{!}{
    \begin{tabular}{c >{\centering\arraybackslash}m{3cm} >{\centering\arraybackslash}m{3.5cm} >{\centering\arraybackslash}m{4cm} >{\centering\arraybackslash}m{3cm} >{\centering\arraybackslash}m{3cm} >{\centering\arraybackslash}m{3cm} c}
        \hline
        Dataset & Task & Scenes & Robot & Robot Morphology & Sensors (D: Depth camera. W: Wrist camera) & Language Annotations & Year\\
        \hline
        Berkeley MVP Data \cite{Radosavovic2022} & Action & Multi-scene (toy kitchen/table), basic motor tasks, 480 trajectories & xArm & Single Arm & RGB, W & Templated & 2022\\
        \hline
        Berkeley RPT Data \cite{Radosavovic2023} & Action & Multi-scene, pick/stack tasks, 960 trajectories & Franka & Single Arm & RGB, W & Templated & 2023\\
        \hline
        KAIST Nonprehensile Objects \cite{kimpre} & Action & 1 scene, non-grasping manipulation, 201 trajectories & Franka & Single Arm & RGB & Natural & 2023\\
        \hline
        QUT Dynamic Grasping \cite{burgess2022dgbench} & Action & 1 scene, dynamic object grasping, 812 trajectories & Franka & Single Arm & RGB, W & None & 2022\\
        \hline
        Stanford MaskVIT Data \cite{gupta2022maskvit} & Action & Multi-scene (container), object pushing/picking, 9.2k trajectories & Sawyer & Single Arm & RGB & None & 2022\\
        \hline
        LSMO Dataset \cite{Osa22} & Action & 1 scene, obstacle avoidance, 50 trajectories & Cobotta & Single Arm & RGB & Natural & 2022\\
        DLR Sara Pour Dataset \cite{padalkar2023guiding} & Action & 1 scene, cup-to-cup pouring, 100 trajectories & DLR SARA & Single Arm & RGB & None & 2023\\
        DLR Sara Grid Clamp Dataset \cite{padalkar2023guided} & Action & 1 scene, grid clamp placement, 100 trajectories & DLR SARA & Single Arm & RGB & None & 2023\\
        \hline
        DLR Wheelchair Shared Control \cite{vogel_edan_2020,quere_shared_2020} & Action & Multi-scene (table/shelf), object grasping, 100 trajectories & DLR EDAN & Single Arm & RGB & Templated & 2020\\
        ASU TableTop Manipulation \cite{zhou2023modularity, zhou2023learning} & Action & 1 scene, object interaction, 110 trajectories & UR5 & Single Arm & RGB & Templated & 2023\\
        Stanford Robocook \cite{shi2023robocook} & Action & Multi-scene (kitchen), dough manipulation, 2.5k trajectories & Franka & Single Arm & RGB, D & Templated & 2023\\
        ETH Agent Affordances \cite{schiavi2023learning} & Action & 1 scene (kitchen), oven/door manipulation, 120 trajectories & Franka & Mobile Manipulator & D & Templated & 2023\\
        Imperial Wrist Cam & Action & Multi-scene (household), object interaction & Sawyer & Single Arm & RGB, W & Natural &\\
        CMU Franka Pick-Insert Data \cite{saxena2023multiresolution} & Action & 1 scene, peg insertion, 520 trajectories & Franka & Single Arm & RGB, W & Templated & 2023\\
        QUT Dexterous Manpulation \cite{ceola2023lhmanip} & Action & Multi-scene (table/kitchen), food serving/tidying, 200 trajectories & Franka & Mobile Manipulator & RGB, W & Natural & 2023\\
        MPI Muscular Proprioception \cite{guist2023robust} & Action & 1 scene (lab), random motion & PAMY2 & Single Arm & - & None & 2023\\
        UIUC D3Field \cite{wang2023d3field} & Action & Multi-scene (household), object arrangement, 196 trajectories & Kinova Gen3 & Single Arm & RGB, D & None & 2023\\
        \hline
        Austin Mutex \cite{shah2023mutex} & Action & Multi-scene (Home Environment), 100+ tasks (pick-place, contact-rich), 1,500 trajectories & Franka & Single Arm & RGB, W & Natural Language (GPT4 + human correction) & 2023 \\
        Berkeley Fanuc Manipulation \cite{fanuc_manipulation2023} & Action & 1 scene, multi-task (drawer opening, object pickup), 415 trajectories & Fanuc Mate & Single Arm & RGB, W & Natural & 2023 \\
        \hline
        CMU Food Manipulation \cite{sawhney2021playing} & Action & 1 scene, 21 food types (slicing/manipulation), 4,200 trajectories & Franka & Single Arm & RGB, W & Templated & 2021 \\
        \hline
    \end{tabular}
    }
\end{table}
\begin{table}[t!]
    \centering
    \label{tab:dataset_3}
    \setlength{\tabcolsep}{3pt}
    \resizebox{1\columnwidth}{!}{
    \begin{tabular}{c >{\centering\arraybackslash}m{3cm} >{\centering\arraybackslash}m{3.5cm} >{\centering\arraybackslash}m{4cm} >{\centering\arraybackslash}m{3cm} >{\centering\arraybackslash}m{3cm} >{\centering\arraybackslash}m{3cm} c}
        \hline
        Dataset & Task & Scenes & Robot & Robot Morphology & Sensors (D: Depth camera. W: Wrist camera) & Language Annotations & Year\\
        \hline
        CMU Play Fusion \cite{chen2023playfusion} & Action & 3 scenes (grill, table setup, sink), complex object interaction, 576 trajectories & Franka & Single Arm & RGB & Natural & 2023 \\
        CMU Stretch \cite{bahl2023affordances, mendonca2023structured} & Action & Multi-scene (kitchen, hallways), navigation + interaction, 135 trajectories & Hello Stretch & Mobile Manipulator & RGB & Templated & 2023 \\
        CoryHall \cite{kahn2018self} & Navigation & 1 scene (office hallways), navigation tasks, 7,328 trajectories & RC Car & Wheeled Robot & RGB & None & 2018 \\
        SACSoN \cite{hirose2023sacson} & Navigation & Multi-scene (office, school), pedestrian-rich navigation, 3,000 trajectories & TurtleBot 2 & Wheeled Robot & RGB, D & None & 2023 \\
        \hline
        RoboVQA & Action & Multi-scene (3 office buildings), long-horizon tasks, 61k trajectories & Google Robot & 3 embodiments & RGB, D & Natural & \\
        \hline
        ALOHA \cite{Zhao2023LearningFB} & Action & 1 scene, dexterous tasks (unwrapping, shoe fitting), 451 trajectories & ViperX Bimanual & Bi-Manual & RGB, W & Templated & 2023 \\
        \hline
        DROID \cite{khazatsky2024droid} & Action & Multi-scene (household), object manipulation, 92k trajectories & Franka & Single Arm & RGB, D, W & Natural & 2024 \\
        ConqHose \cite{ConqHoseManipData} & Action & 1 scene (office), vacuum hose manipulation, 139 trajectories & Spot & Mobile Manipulator & RGB & Natural & 2024 \\
        DobbE \cite{shafiullah2023dobbe} & Action & Multi-scene (household), mobile manipulation, 5k trajectories & Hello Stretch & Mobile Manipulator & RGB, D, W & Natural & 2023 \\
        FMB \cite{luo2024fmb} & Action & 1 scene, multi-object manipulation, 1,804 trajectories & Franka & Single Arm & RGB, D, W & Templated & 2024 \\
        IO-AI Office PicknPlace & Action & Multi-scene (office), pick-and-place tasks, 3,847 trajectories & Human & Human & RGB, D, W & Templated & \\
        \hline
        MimicPlay \cite{wang2023mimicplay} & Action & Multi-scene, imitation tasks, 378 trajectories & Franka & Single Arm & RGB & None & 2023 \\
        \hline
        MobileALOHA \cite{fu2024mobile} & Action & Multi-scene (household), mobile dexterity tasks, 276 trajectories & Mobile ALOHA & Mobile Manipulator & RGB & Templated & 2024 \\
        \hline
        RoboSet \cite{bharadhwaj2023roboagent} & Action & Multi-scene, 18k tasks (pick, stack), 18k trajectories & Franka & Single Arm & RGB, D, W & Natural & 2023 \\
        \hline
        TidyBot \cite{wu2023tidybot} & Action & 1 scene (household), object arrangement, 24 trajectories & TidyBot & Mobile Manipulator & - & Text-based placements & 2023 \\
        \hline
    \end{tabular}
    }
\end{table}
\begin{table}[t!]
    \centering
    \label{tab:dataset_4}
    \setlength{\tabcolsep}{3pt}
    \resizebox{1\columnwidth}{!}{
    \begin{tabular}{c >{\centering\arraybackslash}m{3cm} >{\centering\arraybackslash}m{3.5cm} >{\centering\arraybackslash}m{4cm} >{\centering\arraybackslash}m{3cm} >{\centering\arraybackslash}m{3cm} >{\centering\arraybackslash}m{3cm} c}
        \hline
        Dataset & Task & Scenes & Robot & Robot Morphology & Sensors (D: Depth camera. W: Wrist camera) & Language Annotations & Year\\
        \hline
        VIMA \cite{jiang2023vima} & Action & Multi-scene, multimodal tasks, 660k trajectories & UR5 & Single Arm & RGB & Multimodal templated & 2023 \\
        \hline
        SPOC \cite{spoc2023} & Action & Multi-scene (household), LLM-augmented tasks, 233k trajectories & Hello Stretch & Single Arm & RGB, D, W & Scripted + LLM & 2023 \\
        \hline
        Plex RoboSuite \cite{thomas2023plex} & Action & 1 scene, object interaction, 450 trajectories & Franka & Single Arm & RGB, D, W & None & 2023 \\
        \hline
    \end{tabular}
    }
\end{table}
\section{Challenges and Future Directions}
\label{sec:CFD}
Although the development of EMLMs has surged, it still faces numerous challenges. However, it also presents exciting and valuable avenues for future exploration. This section outlines the current challenges and highlights potential future research directions for the development of EMLMs. The discussion is organized into several key areas: technological challenges, data and annotation issues, and ethical and application-related concerns.

\subsection{Technological Challenges}
Cross-modal Alignment: Despite significant advances in multimodal models, achieving precise and efficient alignment across different modalities—such as vision, language, and motion—remains a fundamental challenge. Developing methods to robustly fuse and align these modalities in real-time, particularly for embodied tasks, is a critical research focus. For example, both the current visual-language model, ReKep \cite{huang2024rekep}, and the Vision-Audio model, SoundSpaces \cite{chen2020soundspaces}, depend on effective alignment of data from diverse modalities. In the absence of proper alignment, the accuracy and efficiency of the response are likely to degrade.

Computational Resources and Efficiency: EMLMs demand significant computational resources and storage. A key challenge is to improve computational efficiency, minimizing energy consumption, and optimizing inference speed while preserving high performance. Advances in model compression, distributed computing, and hardware acceleration will be crucial in addressing these challenges. At present, most models have vast numbers of parameters, and both training and inference processes rely on high-performance GPUs, which are time-intensive and expensive. However, Openvla \cite{kim2024openvla} has introduced an approach where a model with only 7 billion parameters can perform a wide range of tasks. This efficiency is realized when the input consists of visual and language data. However, when additional modalities such as LiDAR, audio, pressure, GPS, and other multimodal inputs are incorporated to tackle more complex tasks, the model size, response time, and associated costs tend to increase significantly.

Generalization Across Domains: While multimodal models have demonstrated impressive performance on specific benchmarks or within particular domains, their ability to generalize across diverse contexts or tasks remains limited. Researchers must explore methods to enhance the transferability and adaptability of these models for real-world applications. For instance, current embodied large models are typically categorized into perception models, such as the GPT series, interaction models like 3D-VLA \cite{zhen20243d}, and navigation models such as SG-Nav \cite{yin2024sg}. The scope of tasks these models can address is relatively fixed, and their generalization ability remains suboptimal.

Handling Temporal and Sequential Information: Embodied models must manage dynamic, real-time data and sequential interactions, presenting a significant challenge in processing continuous actions, environmental events, and the temporal dependencies between perception, reasoning, and movement. In the field of interaction, models are typically categorized into short-horizon action policies, such as R3M \cite{R3M}, and long-horizon action policies, like Palm-e \cite{driess2023palm}. However, in the domain of navigation, there is a lack of models designed for long-term continuous navigation.

\subsection{Data and Annotation Issues}
Diversity and Quality of datasets: Existing datasets for embodied multimodal tasks are often limited in terms of diversity, scale, and quality. The shortage of high-quality, real-world datasets that capture complex, multimodal interactions in dynamic environments hinders effective model training. Future efforts should prioritize the development of larger, more diverse, and better-annotated datasets to enhance the robustness and generalization of multimodal models. While current large-scale datasets like the Open X-Embodiment dataset \cite{o2024open} and ARIO dataset \cite{wang2024all} have made notable strides, they predominantly focus on perception and interactive tasks, such as household chores and kitchen operations. These tasks alone are insufficient to support the full range of capabilities required for embodied intelligent agents. Furthermore, the majority of sensors in these datasets rely on cameras, which limits real-world perception. To address this, it is crucial to integrate additional multimodal sensors, such as LiDAR, sound sensors, radar, force sensors, and GPS, to improve the breadth of data available.

In terms of datasets, it's essential to incorporate real-world, dynamic data. This is particularly crucial in embodied tasks, such as robotics and autonomous systems, where acquiring data from real-world environments is challenging due to the unpredictable nature of physical surroundings. To ensure the practical applicability of these models in real-world scenarios, they must be trained on data that accurately reflects dynamic, non-static environments.

\subsection{Applications and Ethical Considerations}
Autonomous Driving and Robotics: As embodied multimodal models begin to find applications in autonomous driving, robotics, and human-robot interaction, ensuring their safety, reliability, and ethical compliance is paramount. There is a need to address the challenges of decision-making in real-time, the interpretability of model outputs, and the mitigation of risks in autonomous systems.

Ethical and Bias Issues: Multimodal models may unintentionally inherit biases present in the training data, leading to unfair or discriminatory outcomes. It is crucial to address these ethical concerns by developing methods that ensure fairness, transparency, and accountability in decision-making processes.

\subsection{Future Research Directions}
Cross-modal Pre-training and Fine-tuning: Future research should explore more efficient strategies for cross-modal pre-training and fine-tuning, enabling models to perform well across a range of tasks, from perception to decision-making, without requiring extensive retraining.

Self-supervised Learning: The development of self-supervised learning techniques will be key in reducing reliance on large labeled datasets. By leveraging unlabeled data, models can learn richer representations, making them more adaptable and scalable.

Integration with Multimodal Reinforcement Learning: A promising direction is the integration of multimodal models with reinforcement learning. By combining perception, action, and feedback loops, embodied agents can continuously improve and adapt their behaviors in dynamic, real-world environments.

End-to-end large models: Currently, there are various large models designed for different tasks, such as perception, navigation, and interaction. However, the future development trend is moving towards end-to-end large models, where a single model handles everything—from processing input instructions to executing the final task. This approach simplifies the process and enhances efficiency.

\section{Conclusions}
\label{sec:CON}
In conclusion, EMLMs represent a cutting-edge frontier in AI research, combining language, vision, perception, and action to tackle complex, real-world problems. This review has explored the development of large models in language, vision, and multimodal domains, with a specific focus on how embodied tasks, including perception, navigation, interaction, and simulation, are advancing the field.

The progress made thus far demonstrates the transformative potential of embodied multimodal models across diverse applications, from autonomous systems to robotics. However, significant challenges remain in terms of cross-modal alignment, computational efficiency, generalization, and data acquisition. Furthermore, the ethical implications of deploying such technologies must be carefully considered.

Looking ahead, the field holds immense promise. Advancements in cross-modal pre-training, self-supervised learning, and reinforcement learning will likely drive the next generation of more capable, adaptable, and efficient models. In particular, the integration of these models in real-world, dynamic environments promises to revolutionize fields such as autonomous robotics, virtual agents, and interactive systems.

As the field matures, addressing the technical and ethical challenges will be essential for the responsible and impactful deployment of EMLMs. Continued research and collaboration across disciplines will play a pivotal role in shaping the future of this exciting area of AI.

\section*{Acknowledgments}
This work was supported in part by the National Natural Science Foundation of China (42101445).






\bibliographystyle{elsarticle-num} 
\bibliography{20241128}






\end{document}